\documentclass{article}

\usepackage{microtype}
\usepackage{graphicx}
\usepackage{subcaption}
\usepackage{booktabs} 
\usepackage{booktabs}
\usepackage{multirow}
\usepackage{colortbl}
\usepackage{xcolor}
\usepackage{graphicx}   
\usepackage{adjustbox}
\usepackage{enumitem}
\usepackage{pifont}
\usepackage{tcolorbox}
\usepackage{subcaption}
\definecolor{lightgray}{gray}{0.95}
\definecolor{topone}{RGB}{213, 60, 56}   
\definecolor{toptwo}{RGB}{230, 120, 60}
\usepackage{xcolor}
\usepackage{wrapfig}
\newcommand{\best}[1]{\textbf{\textcolor{orange!80!black}{#1}}}
\newcommand{\second}[1]{\textit{\textcolor{orange!80!black}{#1}}}
\usepackage{hyperref}

\usepackage[accepted]{icml2026}

\usepackage{amsmath}
\usepackage{amssymb}
\usepackage{mathtools}
\usepackage{amsthm}

\usepackage[capitalize,noabbrev]{cleveref}

\theoremstyle{plain}

\theoremstyle{definition}

\theoremstyle{remark}

\usepackage[textsize=tiny]{todonotes}

\icmltitlerunning{FeRA: Frequency–Energy Constrained Routing for Effective Diffusion Adaptation Fine-Tuning}

\begin{document}

\twocolumn[
  \icmltitle{FeRA: Frequency–Energy Constrained Routing for \\Effective Diffusion Adaptation Fine-Tuning}



  \icmlsetsymbol{equal}{*}

  \begin{icmlauthorlist}
    \icmlauthor{Bo Yin}{nus,equal}
    \icmlauthor{Xiaobin Hu}{nus,equal}
    \icmlauthor{Xingyu Zhou}{uestc}
    \icmlauthor{Yu He}{ntu}
    \icmlauthor{Peng-Tao Jiang}{vivo}
    \icmlauthor{Yue Liao}{nus}
    \icmlauthor{Junwei Zhu}{ten}
    \\
    \icmlauthor{Jiangning Zhang}{zju}
    \icmlauthor{Ying Tai}{nju}
    \icmlauthor{Shuicheng Yan}{nus}
  \end{icmlauthorlist}

  \icmlaffiliation{nus}{National University of Singapore}
\icmlaffiliation{uestc}{University of Electronic Science and Technology of China}
\icmlaffiliation{ntu}{Nanyang Technological University}
\icmlaffiliation{vivo}{vivo}
\icmlaffiliation{ten}{Tencent}
\icmlaffiliation{zju}{Zhejiang University}
\icmlaffiliation{nju}{Nanjing University}

  \icmlcorrespondingauthor{Xiaobin Hu}{ben0xiaobin0hu1@nus.edu.sg}

  \icmlkeywords{Machine Learning, ICML}

  \vskip 0.3in
]



\printAffiliationsAndNotice{\icmlEqualContribution}

\begin{abstract}
Diffusion models have achieved remarkable success in generative modeling, yet how to effectively adapting large pretrained models to new tasks remains challenging. 
We revisit the reconstruction behavior of diffusion models during denoising to unveil the underlying frequency–energy mechanism governing this process. 
Building upon this observation, we propose \textbf{FeRA}, a frequency-driven fine-tuning framework that aligns parameter updates with the intrinsic frequency–energy progression of diffusion. 
FeRA establishes a comprehensive frequency–energy framework for effective diffusion adaptation fine-tuning, comprising three synergistic components: \textit{(i)} a compact frequency–energy indicator that characterizes the latent’s bandwise energy distribution, \textit{(ii)} a soft frequency router that adaptively fuses multiple frequency-specific adapter experts, and \textit{(iii)} a frequency–energy consistency regularization that stabilizes diffusion optimization and ensures coherent adaptation across bands.
Routing operates in both training and inference, with inference-time routing dynamically determined by the latent frequency energy.
It integrates seamlessly with adapter-based tuning schemes and generalizes well across diffusion backbones and resolutions. By aligning adaptation with the frequency–energy mechanism, \textbf{FeRA} provides a simple, stable, and compatible paradigm for effective and robust diffusion model adaptation. The code are available: \url{https://github.com/YinBo0927/FeRA.git}.
\end{abstract}

\section{Introduction}
Diffusion models have fundamentally reshaped generative modeling~\cite{ho2020denoising, rombach2022high, hu2020face, ji2025sonic}. Advances in denoising-based likelihood learning, consistency training, and distillation have stabilized optimization and accelerated sampling~\cite{ho2020denoising,song2023consistency,salimans2022progressive,ji2024realtalk,xiaobin2025vtbench}, while the shift from compact U-Nets to cross-scale attention with strong semantic encoders has improved resolution and alignment~\cite{rombach2022high,radford2021learning}. Beyond generic image synthesis, diffusion models now power personalization, editing and controllable generation~\cite{ruiz2023dreambooth,gal2022image,zhang2023adding,poole2022dreamfusion}. Open frameworks such as Stable Diffusion~\cite{rombach2022high} bridge research and deployment, making it feasible and increasingly essential to adapt large pretrained backbones to new tasks without harming generalization. This motivates a central challenge: how to efficiently and reliably adapt powerful diffusion models to diverse downstream scenarios within limited computational and storage budgets, placing parameter-efficient fine-tuning (PEFT)~\cite{hu2022lora,mou2024t2i} at the center of attention.

\begin{figure}[t!]
    \centering
    \includegraphics[width=\linewidth]{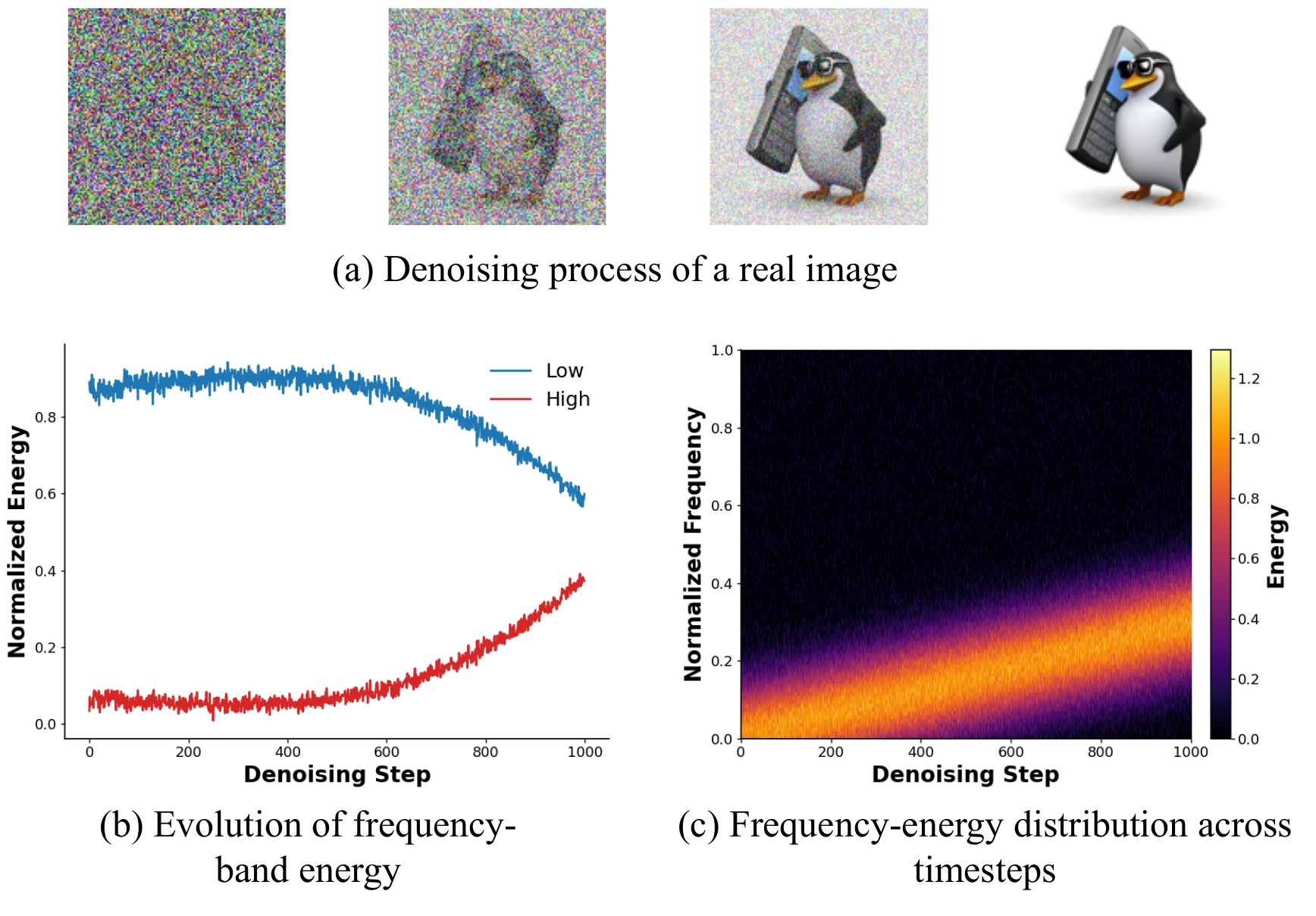}
    \vspace{-6mm}
    \caption{
    Frequency-energy evolution during denoising.
    (a) Visualization of the denoising process. (b) Evolution of frequency-band energies. (c) Frequency-energy distribution across timesteps.
    }
    \label{fig:freq-evo}
    \vspace{-5mm}
\end{figure}

\begin{figure}[t!]
    \centering
    \includegraphics[width=\linewidth]{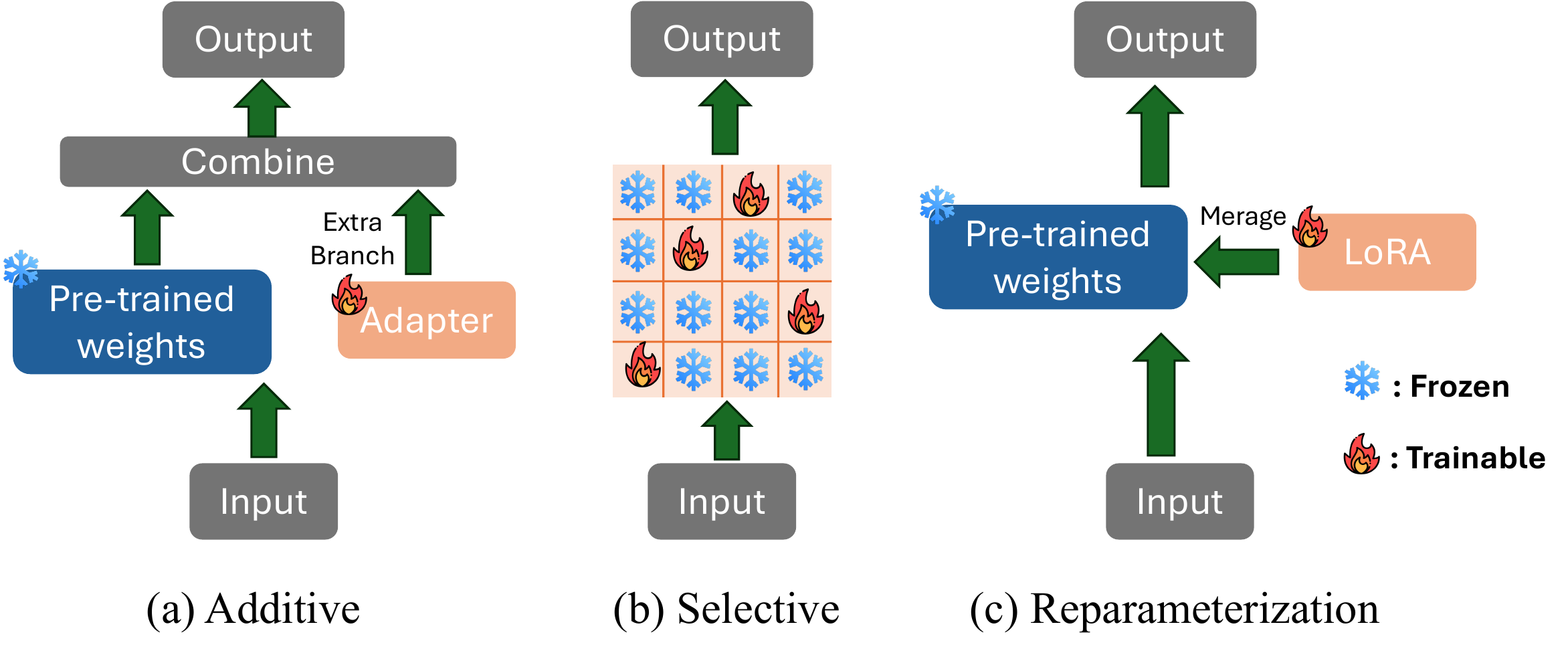}
    \vspace{-5mm}
    \caption{The classical parameter-efficient fine-tuning methods.}
    \label{fig:peft}
    \vspace{-7mm}
\end{figure}

The research trajectory of PEFT has generally followed three main routes in Fig.~\ref{fig:peft}: \textit{\textbf{1)}} \textbf{Additive methods} attach external adapters or side branches, which are simple to train but increase inference overhead~\cite{houlsby2019parameter,mou2024t2i}. \textit{\textbf{2)}} \textbf{Reparameterization methods}, such as low-rank decomposition, embed learnable updates into the original weights for seamless merging at inference~\cite{hu2022lora}. \textit{\textbf{3)}} \textbf{Selective methods}  tune only a small subset of parameters or channels, preserving pretrained priors but increasing implementation complexity~\cite{guo2020parameter}. 
These existing fine-tuning paradigms, including additive, reparameterization, and selective methods, still face inherent limitations. These methods uniformly handle noise across all timesteps. However, diffusion denoising is inherently stage-varying, where the model exhibits distinct noise-signal characteristics over time~\cite{ho2020denoising, nichol2021improved, karras2022elucidating,chen2024find, yang2023diffusion}, as shown in Fig.~\ref{fig:freq-evo}. Consequently, such uniform adaptation fails to align with the intrinsic noise-dependent dynamics of the diffusion process. This raises a fundamental question: \textit{\textbf{Can we design a fine-tuning strategy that adapts to the varying noise conditions throughout the diffusion process rather than enforcing uniform updates across all timesteps?}}

Given the stage-varying nature of diffusion denoising, a dynamic adaptation mechanism across timesteps naturally becomes desirable. Such dynamic capacity can, in principle, be achieved through mixture-of-experts (MoE) mechanisms. Inspired by the success of MoE architectures, recent studies attempt to address this limitation by introducing dynamic routing mechanisms that selectively activate different experts across timesteps or layers. Such designs provide adaptive capacity and improve specialization, yet most routing keys remain discrete and are typically driven by timestep or structural index \textbf{ignoring the physical prior in diffusion process} that often leads to unstable optimization and limited generalization across diffusion backbones. To move beyond these discrete and task-specific schemes, we explore \textit{\textbf{whether the routing process itself can be guided by the continuous pattern observed in diffusion denoising rather than by manually defined timestep indices, thereby enabling fine-tuning to follow the model’s inherent denoising progression.}}

Therefore, we propose \textbf{FeRA}, a frequency-energy-driven framework for parameter-efficient fine-tuning. FeRA employs {frequency experts} that reflect the {frequency-energy} progression observed in diffusion denoising. At each step, the latent representation is decomposed into distinct frequency bands, and their energy proportions are fed into a {frequency router} to produce continuous routing weights. These weights softly blend the frequency experts, enabling the model to adapt smoothly across energy domains while preserving pretrained priors. To further stabilize training, a {frequency-consistency regularization} constrains the discrepancy of update magnitudes between adjacent bands. The unified routing and regularization generalize across model scales and configurations, providing a stable, transferable, and lightweight paradigm for energy-aware fine-tuning. The main contributions of this work are as follows:
\vspace{-4mm}
\begin{itemize}
\item \textbf{Frequency-energy analysis of diffusion denoising.}
We reveal a consistent coarse-to-fine progression that links denoising steps to frequency-energy composition.
\item \textbf{Frequency-driven routing architecture (FeRA).}
We propose a Frequency Energy Indicator to characterize the frequency-energy evolution across timesteps, and leverage it to guide a soft frequency-based expert routing mechanism that replaces discrete timestep hard routing.
\item \textbf{Frequency-energy consistency regularization.}
We introduce a lightweight frequency-energy-based regularization that stabilizes training and enhances the transferability of efficient diffusion adaptation.
\item \textbf{Comprehensive experimental validation.}
We conduct extensive experiments across diverse datasets and diffusion backbones, demonstrating consistent improvements in generation quality and generalization, validating the effectiveness of our frequency-driven design.
\end{itemize}


\section{Related Work}

\subsection{Diffusion Models}
Diffusion models~\cite{ho2020denoising, song2020score, nichol2021improved} have achieved remarkable success in generative modeling by progressively denoising Gaussian noise into structured images.  
Latent diffusion~\cite{rombach2022high} further improves efficiency by operating in a compressed latent space, enabling large-scale text-to-image generation exemplified by Stable Diffusion.  
Subsequent research has extended diffusion models to a wide range of applications, including conditional generation~\cite{zhang2023adding, mou2024t2i, ye2023ip}, controllable synthesis~\cite{song2025omniconsistency, song2025makeanything}, video generation~\cite{blattmann2023stable, guo2023animatediff}, 3D content creation~\cite{poole2022dreamfusion, wang2023score}, and audio-visual modeling~\cite{singer2022make, popov2021grad}.  
Beyond applications, several studies investigate the theoretical and structural properties of diffusion processes, such as noise scheduling~\cite{karras2022elucidating, san2021noise}, consistency training~\cite{song2023consistency}, and frequency-domain analysis~\cite{li2023zero}.  
These works collectively reveal that diffusion denoising proceeds in a coarse-to-fine manner, where low-frequency structures form before high-frequency details, providing new insights for enhancing generative quality and interpretability.

\subsection{PEFT for Generative Models}
Parameter-efficient fine-tuning (PEFT) techniques have become essential for adapting large diffusion models to new concepts or domains without full retraining~\cite{liu2022p, cao2025task}.  
Early approaches such as Textual Inversion~\cite{gal2022image} and DreamBooth~\cite{ruiz2023dreambooth} personalize models by learning small embeddings or tuning only a subset of parameters.  
LoRA~\cite{hu2022lora} introduces low-rank adaptation and has been widely adopted in diffusion frameworks for its efficiency and scalability~\cite{zhang2023adalora}.  
Extensions including ControlNet~\cite{zhang2023adding}, T2I-Adapter~\cite{mou2024t2i}, and IP-Adapter~\cite{ye2023ip} integrate auxiliary networks to enhance conditional control, while adapter fusion and expert routing~\cite{li2025uni, zhu2024task} further improve flexibility and robustness.  
More recent studies investigate unified and task-aware PEFT architectures~\cite{hu2025high, yin2025don}, reflecting a growing interest in scalable and modular adaptation for large generative models.  
Despite differences in structure and application, these methods share a common objective: to enable controllable and data-efficient fine-tuning, yet they implicitly assume isotropic denoising dynamics and overlook the timestep-wise anisotropy of diffusion reconstruction.

\subsection{Mixture of Expert}
Mixture-of-Experts (MoE) architectures introduce conditional computation by activating different parameter subsets based on input features or contextual cues~\cite{shazeer2017outrageously, fedus2022switch, valadarsky2017learning}. This paradigm improves model capacity and specialization without linearly increasing inference cost. Recent works extend MoE to vision and generative tasks, where expert routing is driven by spatial location, semantic content, or timestep signals~\cite{ganjdanesh2024mixture}. In diffusion models, MoE-based adapters or routers have been explored to decouple timestep-dependent behaviors~\cite{liu2024routers, zhu2024task, park2023denoising, lee2024multi, park2024switch, fei2024scaling}, allowing distinct experts to handle different denoising stages. However, most of these methods rely on discrete timestep gating, which introduces hard boundaries and unstable expert activation during training.

\section{Frequency-Energy in Denoising}
\label{fe}

A diffusion model gradually transforms Gaussian noise into a structured image through iterative denoising~\cite{ho2020denoising, yu2025dmfft, tivnan2025fourier, xu2020learning}. As illustrated in Fig.~\ref{fig:freq-evo}(a), this process exhibits a clear visual transition from chaotic noise to coherent structure. To analyze this progression quantitatively, we compute the 2D Fourier amplitude spectrum \(A_t(f)\) of each intermediate image \(x_t\). By integrating spectral energy within predefined low- and high-frequency bands, we obtain their normalized evolution over timesteps (Fig.~\ref{fig:freq-evo}(b)). The results reveal a consistent shift of energy dominance from low to high frequencies as denoising proceeds. The frequency-energy distribution in Fig.~\ref{fig:freq-evo}(c) further confirms that high-frequency components only become prominent in later stages, indicating that diffusion reconstruction progressively shifts energy from low to high frequencies

This behavior can be explained through the frequency-dependent signal-to-noise ratio (SNR) of the diffusion process~\cite{arora2024low}. Given the forward formulation \(x_t = \sqrt{\alpha_t}\,x_0 + \sqrt{1-\alpha_t}\,\epsilon\), its Fourier-domain representation is
\begin{equation}
\hat{x}_t(f) = \sqrt{\alpha_t}\,\hat{x}_0(f) + \sqrt{1-\alpha_t}\,\hat{\epsilon}(f).
\end{equation}
The per-frequency SNR is defined as
\begin{equation}
\mathrm{SNR}_t(f) = \frac{\alpha_t |\hat{x}_0(f)|^2}{(1 - \alpha_t)\, \mathbb{E}[|\hat{\epsilon}(f)|^2]}
\;\propto\; \frac{\alpha_t}{(1-\alpha_t)f^{\gamma}},
\end{equation}
since natural images approximately follow a power-law spectrum \( |\hat{x}_0(f)|^2 \propto 1/f^{\gamma} \) with \( \gamma \approx 2 \)~\cite{field1987relations, ruderman1993statistics}, implying that low frequencies carry substantially higher energy than high frequencies and we note that the denominator $\mathbb{E}[|\hat{\epsilon}(f)|^2]$ remains nearly constant across frequencies, so the frequency dependence of $\mathrm{SNR}_t(f)$ primarily arises from the signal term. 
This relation indicates that both smaller \(\alpha_t\) and higher \(f\) lead to rapidly diminishing SNR.
When $\alpha_t$ is small at early timesteps, the signal term $|\hat{x}_0(f)|^2$ becomes negligible compared to the noise power. 
This effect is particularly pronounced at high frequencies, where the intrinsic decay $|\hat{x}_0(f)|^2 \propto 1/f^\gamma$ further suppresses the signal energy.
As a result, the observed spectrum $\hat{x}_t(f)$ in these bands is effectively noise-dominated and contains almost no recoverable structure. 
In contrast, low-frequency components maintain substantially higher SNR, allowing coarse spatial layouts to remain statistically discernible even in the early stages of denoising. 
As $\alpha_t$ increases over time, the effective SNR of high-frequency bands gradually improves, enabling the model to recover fine-grained details in later denoising steps.

Since the VAE encoder $\mathcal{E}$ is locally approximately linear within the manifold of natural images~\cite{bengio2013representation, kingma2019introduction, rombach2022high}, its latent representation $z_t=\mathcal{E}(x_t)$ satisfies $\hat{z}_t(f)\approx H(f)\hat{x}_t(f)$, where $H(f)$ denotes the encoder’s local frequency response that approximately preserves the spectral structure of natural images, leading to a matching latent SNR, $\mathrm{SNR}^{(z)}_t(f)\approx\mathrm{SNR}_t(f)$. Hence, the same evolution persists within the latent domain, where we later define our frequency indicators and routing strategy, ensuring that the model’s adaptive behavior remains aligned with the underlying spectral progression throughout the entire denoising trajectory.

\section{Frequency-Energy Constrained Routing}
\begin{figure*}[t!]
    \centering
    \includegraphics[width=0.90\linewidth]{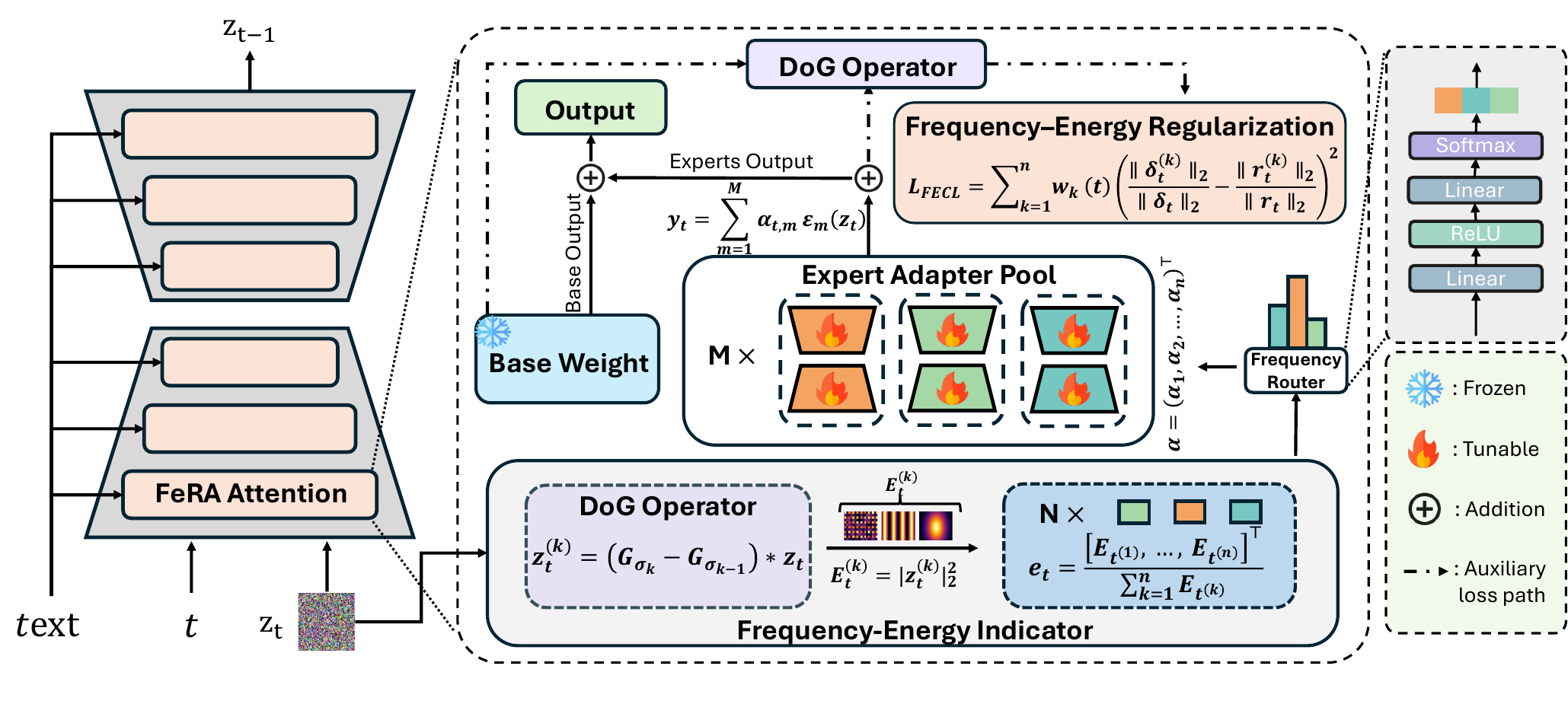}
    \vspace{-5mm}
    \caption{
        Overview of the FeRA framework. The \textbf{Frequency-Energy Indicator (FEI)} extracted by DoG operators guides a \textbf{Soft Frequency Router} to adaptively blend multiple LoRA experts. A \textbf{Frequency-Energy Consistency Loss (FECL)} further regularizes the spectral alignment between correction and residual during fine-tuning.}
    \label{fig:method-overview}
    \vspace{-5mm}
\end{figure*}

Motivated by the frequency-energy analysis in Section~\ref{fe}, we propose \textbf{FeRA}, a parameter-efficient fine-tuning framework that introduces frequency-energy-aware mechanisms at both architectural and training levels. As illustrated in Fig.~\ref{fig:method-overview}, FeRA first employs a Difference-of-Gaussians (DoG)~\cite{lowe2004distinctive} operator to extract the relative energy distribution across different frequency bands, forming a compact \textit{Frequency-Energy Indicator (FEI)} that characterizes the latent’s frequency-energy state. The FEI is then fed into a \textit{Soft Frequency Router}, which adaptively blends multiple LoRA experts according to the current frequency-energy composition, providing a continuous and interpretable alternative to timestep-based hard routing. To further ensure consistency during training, FeRA incorporates a \textit{Frequency-Energy Consistency Loss (FECL)} that aligns the denoising trajectory with the inherent frequency-energy evolution. We also give the theoretical analysis of frequency-energy routing compared with timestep routing in Appendix~\ref{sec:theory_formulation} and discuss the potential applicability of our method to non-natural images in the Appendix~\ref{sec:non_natural_generalization}.

\subsection{Frequency-Energy Indicator (FEI)}
\label{sec:fei}
We define a simple descriptor that summarizes the frequency-energy of the latent feature $z_t\!\in\!\mathbb{R}^{C\times H\times W}$ at denoising step $t$. Let $G_{\sigma}$ be a Gaussian blur with standard deviation $\sigma$ measured in latent pixels. In the frequency domain with radial frequency $\rho$, its response is $\widehat{G}_{\sigma}(\rho) = e^{-2\pi^{2}\sigma^{2}\rho^{2}}$. A Difference-of-Gaussians (DoG) $D_{\sigma_i,\sigma_j}=G_{\sigma_i}-G_{\sigma_j}$ is therefore a band-pass filter with response $e^{-2\pi^2\sigma_i^2\rho^2}-e^{-2\pi^2\sigma_j^2\rho^2}$, which is small near $\rho=0$ and at high $\rho$, and peaks at a middle range. Using $n$ Gaussian kernels with scales $\{\sigma_1, \dots, \sigma_n\}$ in a geometric progression (we use $\sigma_k = \kappa \cdot 2^{k-1}$ for $k=1,\dots,n$, with $\kappa = \min(H, W)/128$, implicitly corresponding to increasing frequency cutoffs), we construct $n$ frequency bands. The filtered components are computed as follows:
\begin{equation}
z_t^{(k)} = 
\begin{cases}
G_{\sigma_k} * z_t &\text{if } k=1, \\
(G_{\sigma_k} - G_{\sigma_{k-1}}) * z_t &\text{if } 1 < k < n, \\
(\mathbb{I} - G_{\sigma_{n-1}}) * z_t &\text{if } k = n,
\end{cases}
\end{equation}
where $G_{\sigma_k}$ denotes a Gaussian blur with standard deviation $\sigma_k$, and $*$ is depth-wise convolution.
We measure the energy of each band as:
{\small
\begin{equation}
E_t^{(k)} = \|z_t^{(k)}\|_2^2 =
\sum_{c=1}^{C} \sum_{x=1}^{H} \sum_{y=1}^{W}
\left( z_{t,c}^{(k)}(x,y) \right)^2,\quad k = 1,\dots,n.
\label{eq:fei-energy-n}
\end{equation}
}
By Parseval’s theorem~\cite{kwakernaak1991modern}, the spatial-domain per-band energy $E_t^{(k)}$ equals its frequency-domain counterpart. Hence, the computed per-band quantity $E_t^{(k)}$ is exactly the frequency-domain energy of $z_t^{(k)}$.
With the dyadic spacing above, the filters exhibit minimal overlap and cover the full frequency spectrum, such that $\sum_{k=1}^n E_t^{(k)} \approx \|z_t\|_2^2$.
We define the \textbf{Frequency-Energy Indicator (FEI)} as the normalized energy vector:
\begin{equation}
\mathbf{e}_t = \frac{[E_t^{(1)},\, \dots,\, E_t^{(n)}]^\top}{\sum_{k=1}^n E_t^{(k)}} \in \mathbb{R}^n,
\label{eq:fei-def-n}
\end{equation}
Which lies on the probability simplex and is invariant to global rescaling of $z_t$, offering a stable descriptor. In Appendix~\ref{sec:fei_analysis}, we further analysis that FEI is not a proxy of timestep.

\subsection{Soft Frequency Router}
\label{sfr}
Building upon FEI, we introduce a \textbf{Frequency-Aware Routing} mechanism that dynamically blends multiple LoRA experts according to the latent’s spectral state. Specifically, the FEI is projected through a lightweight MLP router $g_\phi$ to produce $M$ routing logits and weights:
\begin{equation}
\boldsymbol{\alpha}_t = \mathrm{softmax}\!\left(\frac{g_\phi(\mathbf{e}_t)}{\tau}\right)\in\mathbb{R}^M
\end{equation}
where $\mathbf{e}_t$ is FEI, $\tau$ controls routing softness, and $\alpha_{t,m}$ is the weight assigned to expert $\mathcal{E}_k$. The routed adapter output is the weighted mixture
\begin{equation}
y_t = \sum_{m=1}^{M} \alpha_{t,m}\,\mathcal{E}_m(z_t)
\end{equation}
enabling smooth transitions across different frequency experts as the energy distribution evolves. This frequency-driven formulation improves continuity and interpretability over timestep-based hard gating. 
In practice we attach the frequency-energy experts to the same layers, use a small two-layer router, and fix $\tau{=}0.7$.

\subsection{Frequency-Energy Consistency Loss (FECL)}
\label{fecl}
While the frequency router adaptively adjusts the contribution of LoRA experts, it does not explicitly regularize the spectral behavior of the latent representation. To enforce a consistent evolution of frequency energy during denoising, we introduce a \textbf{Frequency-Energy Consistency Loss (FECL)} applied in the latent space.

Let the latent predicted by the base model be $z_t^{\text{base}}$ and the one adapted by LoRA be $z_t^{\text{lora}}$. We define the correction and reconstruction errors as
\begin{equation}
\delta_t = z_t^{\text{lora}} - z_t^{\text{base}}, \qquad 
r_t = z_t^{\text{lora}} - z_t,
\end{equation}
where $z_t$ is the ground-truth latent at step $t$. Using the $n$-band decomposition from Sec.~\ref{sec:fei}, we apply DoG filters $D_{\sigma_{k-1}, \sigma_k}$ to obtain bandwise components:
\begin{equation}
\left(\delta_t^{(1)}, \dots, \delta_t^{(n)}\right) = D(\delta_t), \quad
\left(r_t^{(1)}, \dots, r_t^{(n)}\right) = D(r_t),
\end{equation}
where $\delta_t = z_t^{\text{lora}} - z_t^{\text{base}}$ and $r_t = z_t^{\text{lora}} - z_t$ denote the adapter correction and residual error, respectively.
We then align the correction with the residual in the frequency domain via the \textbf{Frequency-Energy Consistency Loss (FECL)}:
\begin{equation}
\mathcal{L}_{\text{FECL}} =
\sum_{k=1}^{n} 
w_k(t)\,\left(
\frac{\|\delta_t^{(k)}\|_2}{\|\delta_t\|_2}
-
\frac{\|r_t^{(k)}\|_2}{\|r_t\|_2}
\right)^{\!2}
\end{equation}
where $w_k(t)$ are frequency-band weights derived from the current FEI (we use $w_k(t)=\tilde{E}_t^{\,(k)}/\sum_j\tilde{E}_t^{\,(j)}$). The full objective is
\begin{equation}
\mathcal{L}=\mathcal{L}_{\text{denoise}}+\lambda_{\text{f}}\mathcal{L}_{\text{FECL}}.
\end{equation}
Minimizing $\mathcal{L}_{\text{FECL}}$ enforces \textit{frequency-energy alignment} between the adapter correction and the residual across frequencies, concentrating updates where residual energy is present and suppressing updates where it is negligible.

\section{Experiment}
\begin{table*}[t!]
\small
\renewcommand{\arraystretch}{0.93}
\setlength{\tabcolsep}{2.3pt}
\centering
\caption{Comparison with different parameter-efficient fine-tuning methods on Stable Diffusion 2.0, 3.0 and FLUX.1. \best{Orange bold} = best \second{Light orange italic} = second best. \textbf{\textit{Notice:} the training parameter is same for fair comparison by controlling the rank of FeRA.}}
\vspace{-2mm}
\resizebox{\linewidth}{!}{
\begin{tabular}{c|c|c|ccc|ccc|ccc|ccc|ccc}
\toprule
\multirow{2}{*}{Backbone} & \multirow{2}{*}{Params} & \multirow{2}{*}{Method}
& \multicolumn{3}{c}{Barbie}
& \multicolumn{3}{c}{Cyberpunk}
& \multicolumn{3}{c}{Expedition}
& \multicolumn{3}{c}{Hornify}
& \multicolumn{3}{c}{Elementfire} \\
& & & CLIP $\uparrow$ & FID $\downarrow$ & Style $\uparrow$
& CLIP $\uparrow$ & FID $\downarrow$ & Style $\uparrow$
& CLIP $\uparrow$ & FID $\downarrow$ & Style $\uparrow$
& CLIP $\uparrow$ & FID $\downarrow$ & Style $\uparrow$
& CLIP $\uparrow$ & FID $\downarrow$ & Style $\uparrow$ \\
\midrule

\multirow{16}{*}{SD 2.0}
& \multirow{5}{*}{5M}
& LoRA & \second{35.67} & 213.42 & 8.46 & 33.23 & 178.17 & 8.33 & 31.85 & 164.21 & 8.54 & 32.31 & 175.16 & 8.57 & 32.18 & 200.45 & 8.60 \\
& & DoRA & 35.61 & 211.63 & 8.49 & 33.26 & 176.48 & 8.34 & 31.86 & 162.93 & 8.55 & 32.33 & 173.12 & 8.58 & 32.20 & 198.55 & 8.61 \\
& & AdaLoRA & 35.61 & 212.45 & 8.48 & 33.26 & 177.30 & \second{8.37} & \second{31.86} & 163.61 & 8.56 & \second{32.34} & 174.12 & 8.59 & 32.12 & 199.50 & 8.62 \\
& & SaRA & 35.53 & \second{207.12} & \second{8.51} & \second{33.28} & \second{170.81} & 8.36 & 31.88 & \second{160.37} & \second{8.58} & 32.26 & \second{170.28} & \second{8.60} & \second{32.36} & \second{190.28} & \second{8.64} \\
& &  \textbf{FeRA (Ours)} & \best{36.63} & \best{201.94} & \best{8.52} & \best{33.38} & \best{166.54} & \best{8.39} & \best{33.98} & \best{156.36} & \best{8.61} & \best{32.46} & \best{166.02} & \best{8.63} & \best{33.32} & \best{189.94} & \best{8.65} \\
\cmidrule(lr){2-18}
& \multirow{5}{*}{20M}
& LoRA & \best{35.78} & 198.95 & \second{8.74} & 33.35 & 165.17 & 8.45 & 31.97 & 152.17 & 7.57 & 32.44 & 162.91 & 8.53 & 32.30 & 186.24 & 8.22 \\
& & DoRA & 35.72 & 197.12 & 8.15 & 33.38 & 164.31 & 8.52 & 31.98 & 151.52 & 8.03 & 32.46 & 160.95 & 8.15 & 32.24 & 184.46 & 8.62 \\
& & AdaLoRA & 35.72 & 197.49 & 8.31 & 33.38 & 164.18 & \second{8.76} & 31.98 & 152.12 & 8.17 & \second{32.46} & 161.88 & \second{8.56} & 32.24 & 185.52 & 8.64 \\
& & SaRA & 35.65 & \second{192.36} & 8.44 & \second{33.40} & \second{158.48} & 8.34 & \second{32.00} & \second{149.13} & \second{8.54} & 32.39 & \second{158.24} & 7.99 & \second{32.48} & \best{177.09} & \second{8.76} \\
& &  \textbf{FeRA (Ours)} & \second{35.74} & \best{187.83} & \best{8.82} & \best{33.50} & \best{154.19} & \best{8.78} & \best{32.12} & \best{145.64} & \best{8.55} & \best{32.58} & \best{154.14} & \best{8.89} & \best{33.44} & \second{178.51} & \best{8.87} \\
\cmidrule(lr){2-18}
& \multirow{5}{*}{50M}
& LoRA & \second{35.80} & 200.12 & 8.16 & 33.37 & 166.83 & \second{8.73} & 31.98 & 151.92 & 8.27 & 32.45 & 162.15 & 8.19 & 32.31 & 185.74 & \second{8.44} \\
& & DoRA & 35.74 & 198.33 & 8.00 & 33.40 & 165.24 & 7.83 & 31.99 & 150.47 & 8.67 & \second{32.47} & 160.21 & 8.16 & 32.25 & 183.96 & 8.06 \\
& & AdaLoRA & 35.74 & 199.31 & \second{8.95} & 33.40 & 165.91 & 8.18 & \second{31.99} & 151.13 & 8.35 & 32.47 & 161.00 & 8.14 & 32.25 & 184.84 & 8.41 \\
& & SaRA & 35.67 & \second{193.18} & 8.47 & \second{33.42} & \second{160.03} & 7.92 & 32.01 & \second{148.33} & \second{8.71} & 32.40 & \second{157.24} & \second{8.55} & \second{32.49} & \second{176.25} & 7.86 \\
& &  \textbf{FeRA (Ours)} & \best{36.76} & \best{189.20} & \best{8.98} & \best{33.52} & \best{156.31} & \best{8.75} & \best{32.31} & \best{144.65} & \best{8.83} & \best{32.59} & \best{153.36} & \best{9.06} & \best{33.45} & \best{175.72} & \best{8.77} \\
\cmidrule(lr){2-18}
& \multirow{1}{*}{866M}
& Full-Tuning
& 35.70 & 191.00 & 8.95
& 33.45 & 158.60 & 8.72
& 32.25 & 146.20 & 8.80
& 32.52 & 155.10 & 9.02
& 32.38 & 179.90 & 8.74 \\
\midrule

\multirow{16}{*}{SD 3.0}
& \multirow{5}{*}{5M}
& LoRA & 36.05 & 189.08 & 8.41 & 33.67 & 163.46 & 8.30 & 31.61 & 147.65 & 8.49 & 32.21 & \second{170.31} & 8.53 & 32.19 & 170.39 & 8.54 \\
& & DoRA & \best{36.09} & 189.62 & 8.43 & \second{33.72} & 158.02 & 8.29 & 31.60 & 147.18 & 8.52 & 32.22 & 172.16 & 8.55 & 32.20 & \second{169.27} & 8.56 \\
& & AdaLoRA & \second{36.08} & 188.91 & 8.44 & 33.72 & 161.51 & \second{8.31} & 31.61 & 147.43 & 8.54 & 32.22 & \best{170.12} & 8.55 & \second{32.22} & 169.80 & 8.57 \\
& & SaRA & 36.00 & \second{186.50} & \second{8.45} & 33.71 & \second{155.22} & 8.28 & \second{31.64} & \second{141.02} & \second{8.55} & \best{33.15} & 171.27 & \second{8.56} & \best{32.25} & 170.32 & \second{8.58} \\
& &  \textbf{FeRA (Ours)} & \best{36.09} & \best{184.09} & \best{8.48} & \best{33.81} & \best{154.51} & \best{8.33} & \best{31.67} & \best{139.05} & \best{8.57} & \second{32.34} & 174.25 & \best{8.59} & 31.79 & \best{168.23} & \best{8.60} \\
\cmidrule(lr){2-18}
& \multirow{5}{*}{20M}
& LoRA & 36.15 & 176.41 & 8.55 & 33.78 & 152.41 & 7.81 & 31.72 & 137.34 & \second{8.59} & 32.33 & \second{158.35} & 8.27 & 32.30 & 158.93 & 8.36 \\
& & DoRA & \second{36.20} & 176.92 & \second{8.77} & \second{33.83} & 147.55 & \second{8.02} & 31.71 & 136.69 & 7.86 & 32.34 & 160.22 & 8.30 & \second{32.31} & 157.84 & 8.26 \\
& & AdaLoRA & 36.19 & 176.24 & 8.76 & \second{33.83} & 150.27 & 7.90 & 31.71 & 137.12 & 8.52 & 32.34 & 158.42 & 8.57 & 32.31 & 158.31 & \second{8.67} \\
& & SaRA & 36.10 & \second{173.91} & 8.23 & 33.82 & \second{144.59} & 7.79 & \second{31.75} & \second{131.20} & 7.64 & \best{33.27} & 159.13 & \second{8.61} & \best{32.36} & \second{154.31} & 7.79 \\
& &  \textbf{FeRA (Ours)} & \best{37.20} & \best{171.55} & \best{8.78} & \best{33.92} & \best{141.22} & \best{8.54} & \best{31.78} & \best{129.43} & \best{8.60} & \second{32.46} & \best{151.94} & \best{8.83} & 31.90 & \best{150.86} & \best{8.71} \\
\cmidrule(lr){2-18}
& \multirow{5}{*}{50M}
& LoRA & 36.18 & 177.48 & \second{8.73} & 33.80 & 153.13 & \second{7.99} & 31.73 & 136.72 & 8.47 & 32.35 & 157.87 & 7.87 & 32.32 & 158.71 & 8.26 \\
& & DoRA & 36.23 & 178.63 & 8.01 & \second{33.85} & 148.74 & 7.68 & 31.72 & 136.24 & \second{8.59} & 32.36 & 159.64 & \second{8.39} & \second{32.33} & \second{157.53} & 7.61 \\
& & AdaLoRA & \second{36.22} & 177.64 & 8.14 & 33.85 & 151.95 & 7.49 & 31.72 & 136.54 & 8.08 & 32.36 & \best{157.53} & 7.91 & 32.33 & 157.54 & \second{8.52} \\
& & SaRA & 36.13 & \second{175.13} & 8.55 & \second{33.42} & \second{160.03} & 7.92 & 32.01 & \second{148.33} & \second{8.71} & \second{32.40} & \second{157.24} & \second{8.55} & \best{32.38} & 158.12 & 8.13 \\
& &  \textbf{FeRA (Ours)} & \best{37.23} & \best{172.59} & \best{8.81} & \best{33.94} & \best{145.42} & \best{8.42} & \best{32.79} & \best{128.73} & \best{8.65} & \best{33.48} & 161.12 & \best{8.76} & 31.92 & \best{154.95} & \best{8.57} \\
\cmidrule(lr){2-18}
& \multirow{1}{*}{2085M}
& Full-Tuning
& 36.15 & 174.80 & 8.78
& 33.85 & 147.90 & 8.39
& 31.72 & 130.50 & 8.62
& 32.40 & 162.80 & 8.72
& 31.85 & 156.90 & 8.54 \\
\midrule

\multirow{16}{*}{FLUX.1}
& \multirow{5}{*}{5M}
& LoRA   & 36.05 & 175.23 & 8.55 & 33.61 & 146.32 & \second{8.34} & 32.08 & 132.46 & 8.19 & 32.65 & 143.51 & \second{8.66} & 32.52 & 159.35 & 7.82 \\
& & DoRA   & 36.01 & 173.58 & 8.51 & 33.64 & 144.78 & 8.28 & \second{32.10} & 131.34 & \second{8.63} & 32.67 & 141.83 & 8.14 & \second{32.75} & 157.18 & 7.77 \\
& & AdaLoRA& 36.03 & 172.56 & 8.52 & 33.65 & 143.51 & 8.15 & 31.98 & 132.22 & 8.38 & 32.66 & 140.30 & 7.95 & 32.49 & 155.98 & 8.01 \\
& & SaRA   & \second{36.08} & \second{170.12} & \second{8.61} & \second{33.66} & \second{141.59} & 7.93 & \best{32.12} & \second{129.42} & 8.45 & \second{32.70} & \second{139.23} & 7.87 & 32.58 & \best{154.37} & \second{8.10} \\
& &  \textbf{FeRA (Ours)}
          & \best{37.05} & \best{165.87} & \best{8.66} & \best{33.76} & \best{138.05} & \best{8.50} & 32.02 & \best{126.18} & \best{8.72} & \best{33.81} & \best{135.75} & \best{8.75} & \best{33.68} & \second{155.51} & \best{8.77} \\
\cmidrule(lr){2-18}

& \multirow{5}{*}{20M}
& LoRA   & 36.15 & 162.79 & 8.35 & 33.74 & 136.43 & \second{8.45} & 32.20 & 122.19 & \second{8.78} & 32.78 & 133.40 & \second{8.89} & 32.65 & 148.53 & 8.33 \\
& & DoRA   & 36.12 & 158.92 & \second{8.86} & 33.77 & 134.34 & 8.36 & 32.21 & 121.77 & 8.43 & \second{32.85} & \second{128.05} & 8.23 & 32.68 & 146.44 & \second{8.81} \\
& & AdaLoRA& 36.12 & 161.46 & 8.55 & 33.77 & 135.16 & 8.08 & \second{32.23} & 122.73 & 8.60 & 32.83 & 132.63 & 8.20 & 32.68 & 147.62 & 8.22 \\
& & SaRA   & \second{36.19} & \second{156.15} & 8.28 & \second{33.79} & \second{131.42} & 8.21 & \best{32.25} & \second{120.34} & 8.50 & 32.83 & 129.13 & 8.30 & \second{32.71} & \second{143.29} & 8.39 \\
& &  \textbf{FeRA (Ours)}
          & \best{37.16} & \best{153.79} & \best{9.01} & \best{33.89} & \best{128.43} & \best{8.94} & 32.14 & \best{117.83} & \best{8.79} & \best{32.94} & \best{126.11} & \best{9.06} & \best{32.81} & \best{141.15} & \best{9.10} \\
\cmidrule(lr){2-18}

& \multirow{5}{*}{50M}
& LoRA   & 35.17 & 163.76 & 8.43 & 33.75 & 136.85 & 8.39 & 32.21 & 122.23 & \second{8.65} & 32.79 & 132.93 & 8.43 & 32.66 & 148.40 & \second{8.78} \\
& & DoRA   & 36.13 & 159.02 & \second{8.71} & 33.78 & 135.61 & \second{8.97} & 32.14 & 121.15 & 8.02 & \second{32.84} & 131.00 & 8.28 & 32.69 & 145.39 & 8.65 \\
& & AdaLoRA& 36.13 & 162.62 & 8.55 & 33.78 & 135.93 & 8.64 & \second{32.24} & 121.77 & 8.06 & \second{32.84} & 131.48 & 8.92 & 32.69 & 146.67 & 8.34 \\
& & SaRA   & \second{36.20} & \second{156.91} & 8.18 & \second{33.80} & \second{131.28} & 8.19 & \best{33.26} & \second{119.71} & 8.55 & \second{32.84} & \second{128.65} & \second{9.00} & \second{32.72} & \second{143.74} & 8.64 \\
& &  \textbf{FeRA (Ours)}
          & \best{37.17} & \best{154.76} & \best{9.15} & \best{34.90} & \best{129.32} & \best{9.11} & 32.15 & \best{117.22} & \best{8.85} & \best{33.95} & \best{125.36} & \best{9.16} & \best{33.82} & \best{141.21} & \best{9.03} \\
\cmidrule(lr){2-18}

& \multirow{1}{*}{7.8B}
& Full-Tuning
& 36.13 & 165.37 & 8.71
& 33.81 & 137.55 & 8.55
& 32.17 & 125.68 & 8.77
& 32.86 & 135.25 & 8.80
& 32.80 & 153.87 & 8.82 \\

\bottomrule

\end{tabular}}
\vspace{-0.35em}

\label{tab:sd15_sd2_sd3_compact}
\vspace{-0.6em}
\end{table*}

To validate the effectiveness of our method, we conduct experiments on two representative scenarios: downstream dataset fine-tuning and image customization. Specially, we set the number of FEI and LoRA experts to 3, and why we set this number will be discussed in Section~\ref{ab}.
We compare our approach with four state-of-the-art parameter-efficient fine-tuning methods: LoRA~\cite{hu2022lora}, DoRA~\cite{liu2024dora} AdaLoRA~\cite{zhang2023adalora}, and SaRA~\cite{hu2025high}, along with the full-parameter fine-tuning baseline. All models are fine-tuned under same training configurations for comparison. 

\begin{figure*}[!t]
    \centering
    \includegraphics[width=\linewidth]{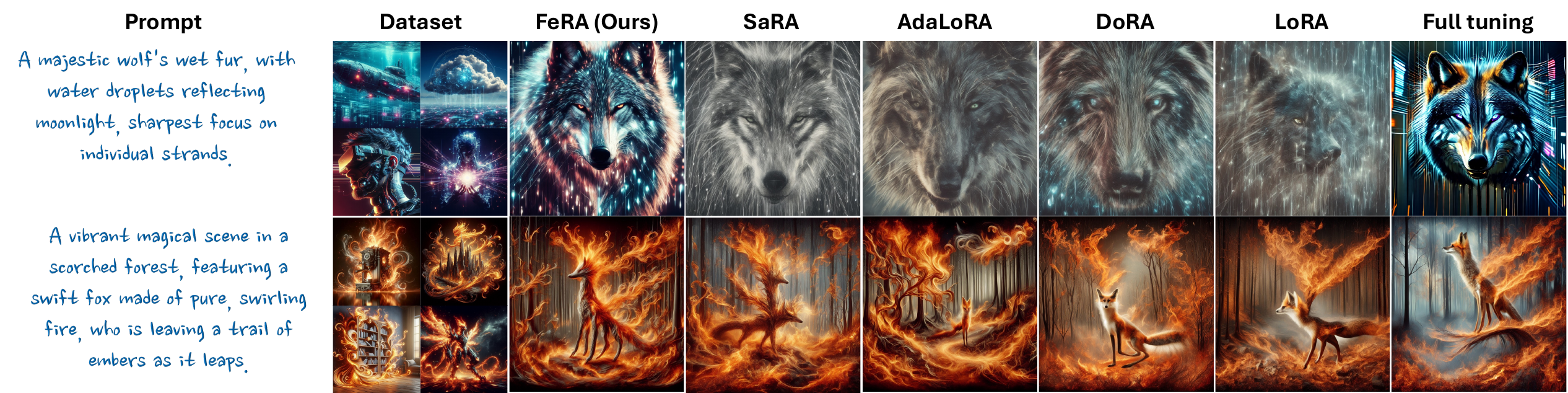}
    \caption{Comparison of the generated images between different PEFT methods.}
    \label{fig:exp_visual}
    \vspace{-4mm}
\end{figure*}

\subsection{Text-to-Image Style Adaptation}
\textbf{Experiment Setting.}
In this experiment, we fine-tune on five widely used CIVITAI style datasets, including Barbie Style, Cyberpunk Style, Elementfire Style, Expedition Style, and Hornify Style. We compare parameter-efficient fine-tuning methods on Stable Diffusion 2.0, 3.0, and FLUX.1 under three trainable parameter budgets of 5M, 20M, and 50M, scaling the rank of FeRA inversely to the number of experts to maintain a fair iso-parameter comparison with baselines. More backbones can be found in Appendix.
Results on additional backbones following the same protocol are provided in the Appendix. Each fine-tuned model is evaluated using three metrics: Fréchet Inception Distance (FID), CLIP Score, and Style Score for perceptual style consistency. The Style Score is obtained through a multimodal large language model (MLLM)-based quantitative evaluation using Qwen2.5-VL-7B-Instruct~\cite{bai2023qwen}. We also give user study analysis in Appendix~\ref{user_study}.

\begin{table*}[!t]
\centering
\caption{
Quantitative comparison between different PEFT methods on image customization.
\textbf{\textcolor{topone}{Red bold}} = best, 
\textit{\textcolor{toptwo}{orange italic}} = second best.
}
\vspace{-2mm}
\small
\setlength{\tabcolsep}{4pt}
\renewcommand{\arraystretch}{1.12}
\resizebox{\linewidth}{!}{%
\begin{tabular}{l|cc|cc|cc|cc|cc}
\toprule
\multirow{2}{*}{Methods} &
\multicolumn{2}{c|}{Dog} &
\multicolumn{2}{c|}{Clock} &
\multicolumn{2}{c|}{Backpack} &
\multicolumn{2}{c|}{Toy Duck} &
\multicolumn{2}{c}{Teapot} \\
\cmidrule(lr){2-11}
 & CLIP-I~$\uparrow$ & CLIP-T~$\uparrow$
 & CLIP-I~$\uparrow$ & CLIP-T~$\uparrow$
 & CLIP-I~$\uparrow$ & CLIP-T~$\uparrow$
 & CLIP-I~$\uparrow$ & CLIP-T~$\uparrow$
 & CLIP-I~$\uparrow$ & CLIP-T~$\uparrow$ \\
\midrule
Dreambooth + Full-tuning & 0.788 & 24.15 & 0.789 & 23.15 & 0.654 & 24.09 & 0.790 & 24.05 & 0.750 & 24.12 \\
Dreambooth + LoRA & \textit{\textcolor{toptwo}{0.895}} & 23.64 & 0.913 & 21.71 & \textit{\textcolor{toptwo}{0.917}} & 25.23 & 0.905 & 23.80 & 0.906 & 23.58 \\
Dreambooth + DoRA & 0.897 & 23.71 & \textit{\textcolor{toptwo}{0.915}} & 21.78 & 0.914 & \textit{\textcolor{toptwo}{25.31}} & \textit{\textcolor{toptwo}{0.907}} & 23.88 & \textit{\textcolor{toptwo}{0.908}} & 23.65 \\
Dreambooth + AdaLoRA & 0.896 & 23.69 & 0.914 & 21.76 & 0.917 & 25.29 & 0.906 & 23.85 & 0.907 & 23.63 \\
Dreambooth + SaRA & 0.790 & \textbf{\textcolor{topone}{25.97}} & 0.887 & \textit{\textcolor{toptwo}{23.51}} & 0.886 & 25.27 & 0.885 & \textbf{\textcolor{topone}{25.50}} & 0.866 & \textit{\textcolor{toptwo}{25.12}} \\
\rowcolor{lightgray}\textbf{Dreambooth + FeRA (Ours)} & \textbf{\textcolor{topone}{0.900}} & \textit{\textcolor{toptwo}{25.95}} & \textbf{\textcolor{topone}{0.920}} & \textbf{\textcolor{topone}{23.63}} & \textbf{\textcolor{topone}{0.925}} & \textbf{\textcolor{topone}{25.35}} & \textbf{\textcolor{topone}{0.910}} & \textit{\textcolor{toptwo}{24.60}} & \textbf{\textcolor{topone}{0.913}} & \textbf{\textcolor{topone}{25.32}} \\
\bottomrule
\end{tabular}%
}
\vspace{-2mm}
\label{tab:clip_results}
\end{table*}

\begin{figure*}[!t]
    \centering
    \vspace{-2mm}
    \includegraphics[width=\linewidth]{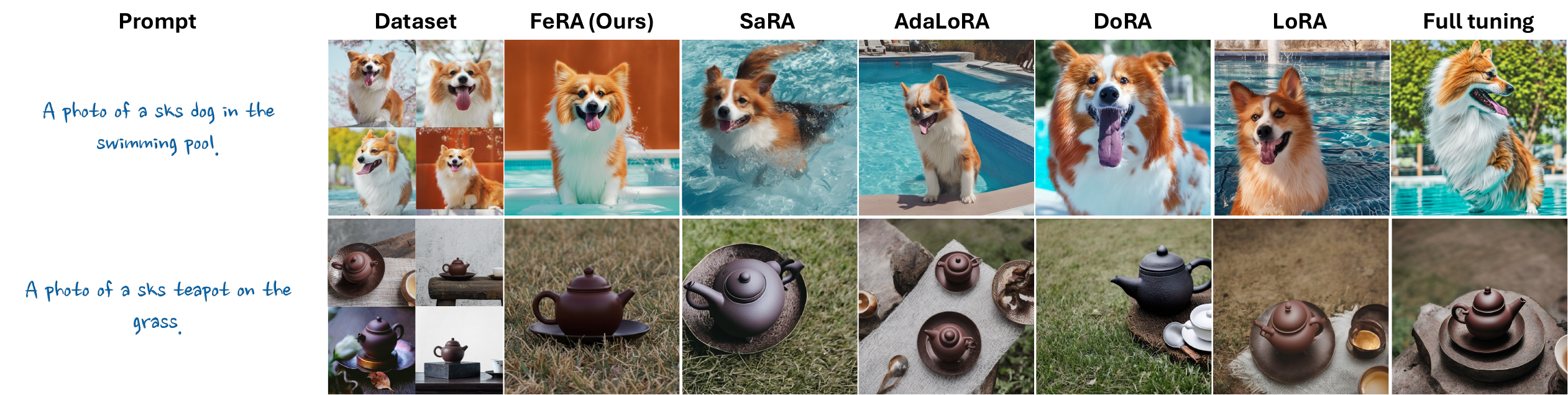}
    \caption{DreamBooth results across PEFT methods. FeRA delivers more consistent identity and cleaner compositions.}
    \label{fig:dr_visual}
    \vspace{-5mm}
\end{figure*}

\noindent
\textbf{Result Analysis.}
The quantitative results are reported in Tab.~\ref{tab:sd15_sd2_sd3_compact}. 
\textit{1).} Across SD~2.0, 3.0 and FLUX1., FeRA consistently ranks among the top-performing methods across most styles, achieving lower FID while maintaining competitive CLIP and Style scores. 
Even under the 5M budget, it attains the lowest FID on several datasets without compromising alignment. 
\textit{2).} As the trainable budget increases to 20M and 50M, FeRA captures finer stylistic details, with FID continuing to improve and Style scores remaining stable, indicating that frequency-energy-aware soft routing scales effectively with model capacity rather than overfitting. 
\textit{3).} The full-parameter baseline is not uniformly dominant under these settings, while FeRA achieves comparable or better performance with substantially fewer parameters. Qualitative comparisons in Fig.~\ref{fig:exp_visual} further shows coherent structure, accurate alignment, and high-fidelity style renderings.

\subsection{Image Customization}
\noindent
\textbf{Experiment Setting.}
We evaluate the proposed FeRA framework on personalized image generation, a representative task that tests a model’s ability to adapt to novel identities or concepts with limited examples. Specifically, we conduct experiments on five representative subject categories, including dog, clock, backpack, toy duck, and teapot. 
Following the standard DreamBooth protocol~\cite{ruiz2023dreambooth}, we fine-tune a pre-trained Stable Diffusion 2.0 model using a small number of instance images (3-5) per subject, binding each identity to a unique rare token. All competing methods are implemented on the same UNet backbone for a fair comparison, including full finetuning, LoRA~\cite{hu2022lora}, DoRA~\cite{liu2024dora}, AdaLoRA~\cite{zhang2023adalora}, and SaRA~\cite{hu2025high}. For quantitative evaluation, we use CLIP-based similarity metrics~\cite{ramesh2022hierarchical} to measure both image-text alignment and visual fidelity, reporting CLIP-IMG and CLIP-Text scores, where higher values indicate better consistency between generated results and target identity descriptions.

\noindent
\textbf{Result Analysis.}
As summarized in Tab.~\ref{tab:clip_results}, FeRA consistently achieves the highest or second-highest scores on both metrics across all categories. For instance, it attains the best CLIP-I values on all five subjects and top CLIP-T results in four of them, surpassing other PEFT methods by a clear margin. LoRA and DoRA perform competitively in CLIP-I but yield noticeably lower CLIP-T, suggesting overfitting toward visual features at the expense of text alignment. SaRA, while producing the highest CLIP-T on dog, suffers a sharp drop in CLIP-I, reflecting poor balance between fidelity and semantic accuracy. Overall, FeRA demonstrates the best trade-off between subject identity preservation and text-image alignment, validating its effectiveness and generalizability in personalized image generation.  
As illustrated in Fig.~\ref{fig:dr_visual}, FeRA produces clearer textures and more accurate colors across all prompts. It preserves structural coherence and fine details, yielding sharper boundaries, more faithful appearance, and more stable rendering across diverse visual conditions, consistently outperforming existing PEFT baselines in visual quality.

\subsection{Ablation Studies}
\label{ab}
\begin{figure*}[t!]
    \centering
    \begin{subfigure}[t]{0.32\linewidth}
        \centering
        \includegraphics[width=\linewidth]{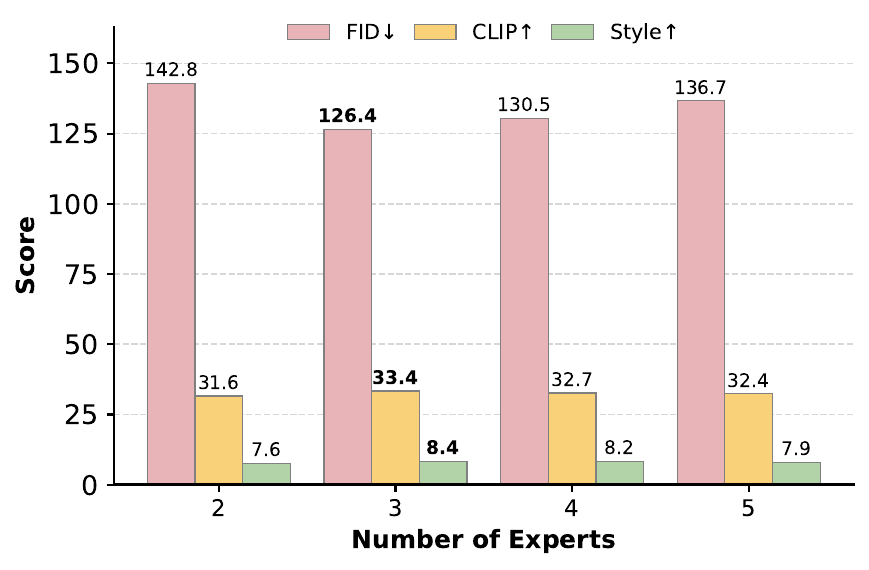}
        \caption{Number of LoRA experts}
    \end{subfigure}
    \begin{subfigure}[t]{0.32\linewidth}
        \centering
        \includegraphics[width=\linewidth]{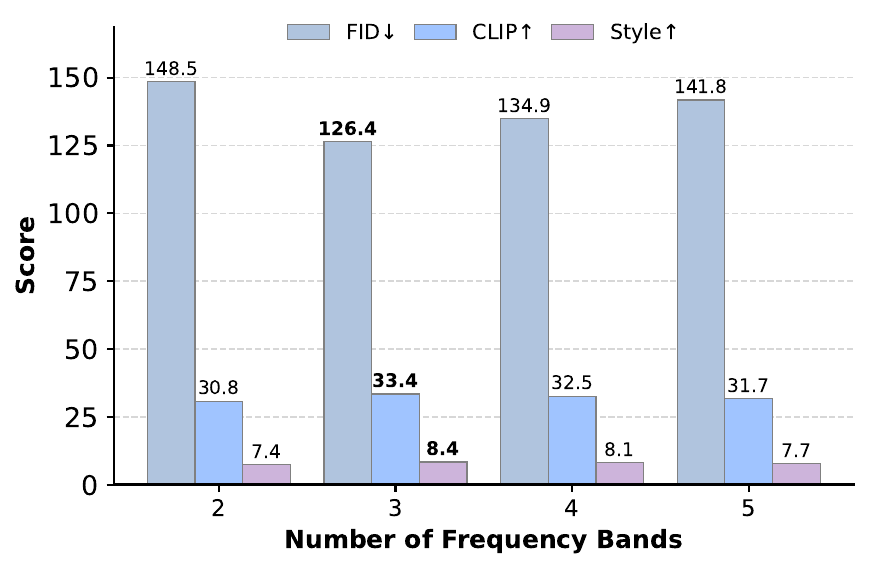}
        \caption{Frequency band decomposition}
    \end{subfigure}
    \begin{subfigure}[t]{0.32\linewidth}
        \centering
        \includegraphics[width=\linewidth]{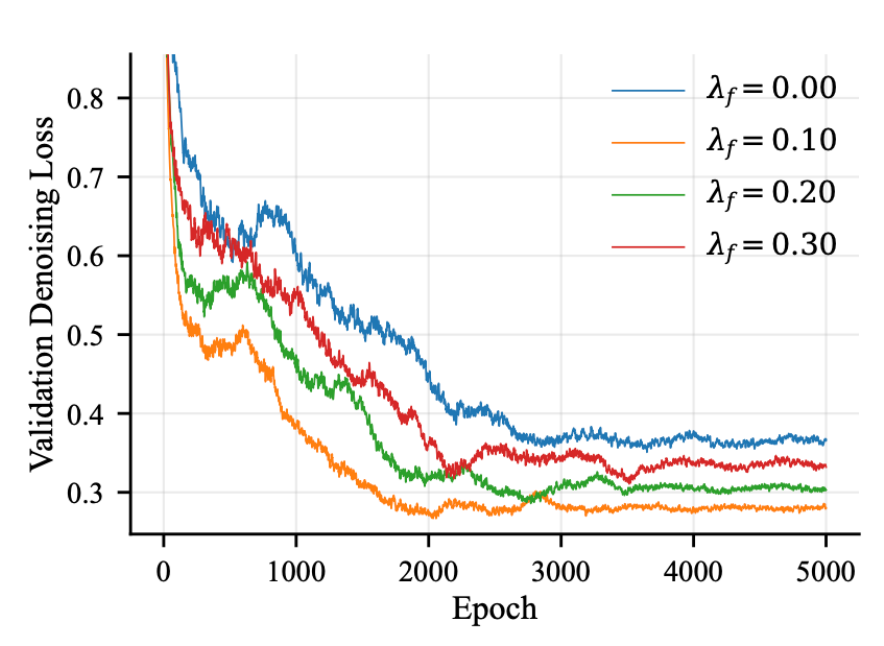}
        \caption{Strengths of FECL.}
    \end{subfigure}
    \vspace{-1mm}
    \caption{Ablation on design factors of FeRA: (a) LoRA expert number, (b) frequency decomposition, and (c) the strengths of FECL.}
    \vspace{-3mm}
    \label{fig:ab}
\end{figure*}
We conduct ablation studies on Stable Diffusion 2.0 using the Cyberpunk dataset to evaluate FeRA's components: FEI, soft router, FECL, and expert/band counts. 

\noindent
\textbf{Timestep-based {v.s.} Frequency-Energy routing.}  Tab.~\ref{tab:ablation-fei-router} reports that replacing {timestep-based routing} with FEI consistently improves FID, CLIP, and Style scores. It indicates that \textit{1)} \textbf{Frequency–energy outperform timestep-based routing} by providing richer and more informative signals; \textit{2)} \textbf{Soft routing outperforms hard routing} due to its smoother and more flexible information aggregation.


\begin{table}[t!]
\centering
\caption{
Ablation on routing configurations. 
“FEI” denotes the proposed frequency–energy indicator, and “Soft Router” refers to the soft routing mechanism. 
Removing FEI falls back to timestep-based routing. “{\color{green!60!black}\ding{51}}” and “{\color{red}\ding{55}}” indicate whether each module is enabled.
}
\vspace{-3mm}
\label{tab:ablation-fei-router}
\setlength{\tabcolsep}{7pt}
\renewcommand{\arraystretch}{1.05}
\begin{tabular}{ccccc}
\toprule
\textbf{FEI} & \textbf{Soft Router} & CLIP$\uparrow$ & FID$\downarrow$ & Style$\uparrow$ \\
\midrule
{\color{red}\ding{55}} & {\color{red}\ding{55}} & 30.12 & 138.50 & 7.42 \\
{\color{green!60!black}\ding{51}} & {\color{red}\ding{55}} & 31.84 & 132.70 & 7.93 \\
{\color{red}\ding{55}} & {\color{green!60!black}\ding{51}} & 31.10 & 134.90 & 7.68 \\
{\color{green!60!black}\ding{51}} & {\color{green!60!black}\ding{51}} & \textbf{32.96} & \textbf{126.40} & \textbf{8.21} \\
\bottomrule
\end{tabular}
\vspace{-3mm}
\end{table}

\begin{figure}[t!]
    \centering
    \includegraphics[width=\linewidth]{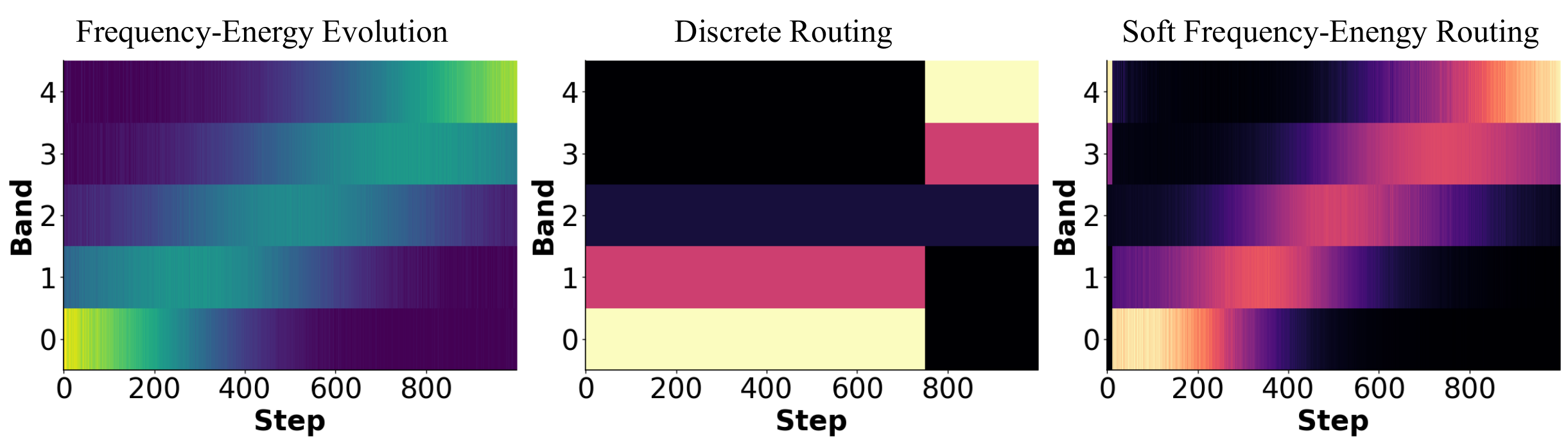}
    \vspace{-5mm}
    \caption{Compared with discrete routing strategy.}
    \vspace{-7mm}
    \label{fig:wan}
\end{figure}


\noindent
\textbf{The number of Experts.}
Under the three-band setting, Figure~\ref{fig:ab} (a) compares different numbers of LoRA experts. Using one expert per band (three in total) achieves the best performance across FID, CLIP, and Style metrics. This confirms that frequency-specialized experts enable more efficient and interpretable adaptation.

\noindent
\textbf{The number of Frequency Band.}
We vary the number of frequency bands used for spectral decomposition in FEI. As shown in Figure~\ref{fig:ab} (b), three bands (low, mid, high) consistently outperform two or more than three. This indicates that a three-band division captures semantic structure while avoiding redundancy and over-fragmentation.

\noindent
\textbf{Frequency-Energy Consistency Loss.}
Figure~\ref{fig:ab} (c) shows the impact of different FECL weights $\lambda_{\mathrm f}\!\in\!\{0,0.1,0.2,0.3\}$. A small weight accelerates and stabilizes convergence, whereas removing FECL slows optimization and larger weights cause oscillation. Moderate regularization achieves the best trade-off between stability and generalization.


\subsection{Our Soft Frequency-Enengy Routing Analysis}
\label{sec:wan_compare}
A representative discrete routing strategy employs a expert MoE design for diffusion denoising: a high-noise expert for early steps that captures global structure, and a low-noise expert for later steps that refines details. The routing is typically governed by a monotonic SNR schedule with a single threshold, resulting in a hard switch between experts. Such a scheme is adopted, for example, in Wan2.2~\cite{wan2025wan}.
Although this threshold-based routing can be extended to multi-LoRA settings for comparison, it inherently enforces a discrete expert transition. To visualize the behavioral difference, we plot the Frequency-Energy Evolution and the corresponding routing weights as heatmaps (Fig.~\ref{fig:wan}). The discrete routing shows an abrupt switch between low- and high-frequency bands, whereas our frequency-aware routing produces a smooth transition that follows the spectral energy migration. This continuity enables finer control and better alignment between the model’s update behavior and the evolving frequency composition during denoising.

\subsection{Inference-Time Comparison}
To quantify inference overhead, we compare LoRA and FeRA using the end-to-end wall-clock latency per generated image across different backbones (Tab.~\ref{tab:fera_vs_lora_time}). The time is measured from the start of the diffusion sampling loop to the final decoded output. All models use identical sampling steps and inference settings for fairness. FeRA adds lightweight routing and frequency-aware operations, incurring only 4-8\% overhead depending on backbone size, while remaining close to LoRA in efficiency and offering better generation quality and controllability.

\begin{table}[t!]
\centering
\caption{Inference-time comparison between LoRA and FeRA across multiple Stable Diffusion backbones.}
\vspace{-2mm}
\small
\resizebox{\columnwidth}{!}{
\begin{tabular}{lccccc}
\toprule
\textbf{Method} & \textbf{SD~1.5} & \textbf{SD~2.0} & \textbf{SD~3.0} & \textbf{SDXL} & \textbf{FLUX.1} \\
\midrule
LoRA (baseline) & $1.00\times$ & $1.00\times$ & $1.00\times$ & $1.00\times$ & $1.00\times$ \\
\rowcolor{lightgray}\textbf{FeRA (ours)} 
& $1.08\times$ 
& $1.08\times$
& $1.07\times$
& $1.05\times$
& $1.04\times$ \\
\bottomrule
\end{tabular}}
\label{tab:fera_vs_lora_time}
\vspace{-8mm}
\end{table}
\vspace{-2mm}
\section{Conclusion}
In this paper, we presented FeRA, a parameter-efficient fine-tuning framework guided by the frequency-energy state of latent representations that utilizes a Frequency-Energy Indicator, Soft Router, and Consistency Loss to align adaptation with the natural frequency hierarchy. Extensive experiments demonstrate that FeRA achieves superior fidelity, alignment, and style customization with minimal parameter overhead. Our findings confirm that frequency-energy awareness offers a robust and interpretable direction for diffusion adaptation.


\section*{Impact Statement}
This paper presents work whose goal is to advance the field of Machine
Learning. There are many potential societal consequences of our work, none
which we feel must be specifically highlighted here.

\nocite{langley00}

\bibliography{example_paper}

@String(ICLR = {Int. Conf. Learn. Represent.})

@String(AAAI = {AAAI})

@String(ICLR  = {ICLR})

@article{hu2022lora,
  title={Lora: Low-rank adaptation of large language models.},
  author={Hu, Edward J and Shen, Yelong and Wallis, Phillip and Allen-Zhu, Zeyuan and Li, Yuanzhi and Wang, Shean and Wang, Lu and Chen, Weizhu and others},
  journal={ICLR},
  volume={1},
  number={2},
  pages={3},
  year={2022}
}

@inproceedings{liu2024dora,
  title={Dora: Weight-decomposed low-rank adaptation},
  author={Liu, Shih-Yang and Wang, Chien-Yi and Yin, Hongxu and Molchanov, Pavlo and Wang, Yu-Chiang Frank and Cheng, Kwang-Ting and Chen, Min-Hung},
  booktitle={Forty-first International Conference on Machine Learning},
  year={2024}
}

@article{zhang2023adalora,
  title={Adalora: Adaptive budget allocation for parameter-efficient fine-tuning},
  author={Zhang, Qingru and Chen, Minshuo and Bukharin, Alexander and Karampatziakis, Nikos and He, Pengcheng and Cheng, Yu and Chen, Weizhu and Zhao, Tuo},
  journal={arXiv preprint arXiv:2303.10512},
  year={2023}
}

@inproceedings{hu2025high,
  title={High-efficient diffusion model fine-tuning with progressive sparse low-rank adaptation},
  author={Hu, Teng and Zhang, Jiangning and Yi, Ran and Huang, Hongrui and Wang, Yabiao and Ma, Lizhuang},
  booktitle={13th International Conference on Learning Representations, ICLR 2025},
  pages={92066--92078},
  year={2025},
  organization={International Conference on Learning Representations, ICLR}
}

@article{ho2020denoising,
  title={Denoising diffusion probabilistic models},
  author={Ho, Jonathan and Jain, Ajay and Abbeel, Pieter},
  journal={Advances in neural information processing systems},
  volume={33},
  pages={6840--6851},
  year={2020}
}

@inproceedings{nichol2021improved,
  title={Improved denoising diffusion probabilistic models},
  author={Nichol, Alexander Quinn and Dhariwal, Prafulla},
  booktitle={International conference on machine learning},
  pages={8162--8171},
  year={2021},
  organization={PMLR}
}

@inproceedings{rombach2022high,
  title={High-resolution image synthesis with latent diffusion models},
  author={Rombach, Robin and Blattmann, Andreas and Lorenz, Dominik and Esser, Patrick and Ommer, Bj{\"o}rn},
  booktitle={Proceedings of the IEEE/CVF conference on computer vision and pattern recognition},
  pages={10684--10695},
  year={2022}
}

@article{song2020score,
  title={Score-based generative modeling through stochastic differential equations},
  author={Song, Yang and Sohl-Dickstein, Jascha and Kingma, Diederik P and Kumar, Abhishek and Ermon, Stefano and Poole, Ben},
  journal={arXiv preprint arXiv:2011.13456},
  year={2020}
}

@inproceedings{zhang2023adding,
  title={Adding conditional control to text-to-image diffusion models},
  author={Zhang, Lvmin and Rao, Anyi and Agrawala, Maneesh},
  booktitle={Proceedings of the IEEE/CVF international conference on computer vision},
  pages={3836--3847},
  year={2023}
}

@inproceedings{mou2024t2i,
  title={T2i-adapter: Learning adapters to dig out more controllable ability for text-to-image diffusion models},
  author={Mou, Chong and Wang, Xintao and Xie, Liangbin and Wu, Yanze and Zhang, Jian and Qi, Zhongang and Shan, Ying},
  booktitle={Proceedings of the AAAI conference on artificial intelligence},
  volume={38},
  number={5},
  pages={4296--4304},
  year={2024}
}

@article{ye2023ip,
  title={Ip-adapter: Text compatible image prompt adapter for text-to-image diffusion models},
  author={Ye, Hu and Zhang, Jun and Liu, Sibo and Han, Xiao and Yang, Wei},
  journal={arXiv preprint arXiv:2308.06721},
  year={2023}
}

@article{blattmann2023stable,
  title={Stable video diffusion: Scaling latent video diffusion models to large datasets},
  author={Blattmann, Andreas and Dockhorn, Tim and Kulal, Sumith and Mendelevitch, Daniel and Kilian, Maciej and Lorenz, Dominik and Levi, Yam and English, Zion and Voleti, Vikram and Letts, Adam and others},
  journal={arXiv preprint arXiv:2311.15127},
  year={2023}
}

@article{guo2023animatediff,
  title={Animatediff: Animate your personalized text-to-image diffusion models without specific tuning},
  author={Guo, Yuwei and Yang, Ceyuan and Rao, Anyi and Liang, Zhengyang and Wang, Yaohui and Qiao, Yu and Agrawala, Maneesh and Lin, Dahua and Dai, Bo},
  journal={arXiv preprint arXiv:2307.04725},
  year={2023}
}

@article{poole2022dreamfusion,
  title={Dreamfusion: Text-to-3d using 2d diffusion},
  author={Poole, Ben and Jain, Ajay and Barron, Jonathan T and Mildenhall, Ben},
  journal={arXiv preprint arXiv:2209.14988},
  year={2022}
}

@inproceedings{wang2023score,
  title={Score jacobian chaining: Lifting pretrained 2d diffusion models for 3d generation},
  author={Wang, Haochen and Du, Xiaodan and Li, Jiahao and Yeh, Raymond A and Shakhnarovich, Greg},
  booktitle={Proceedings of the IEEE/CVF conference on computer vision and pattern recognition},
  pages={12619--12629},
  year={2023}
}

@article{singer2022make,
  title={Make-a-video: Text-to-video generation without text-video data},
  author={Singer, Uriel and Polyak, Adam and Hayes, Thomas and Yin, Xi and An, Jie and Zhang, Songyang and Hu, Qiyuan and Yang, Harry and Ashual, Oron and Gafni, Oran and others},
  journal={arXiv preprint arXiv:2209.14792},
  year={2022}
}

@inproceedings{popov2021grad,
  title={Grad-tts: A diffusion probabilistic model for text-to-speech},
  author={Popov, Vadim and Vovk, Ivan and Gogoryan, Vladimir and Sadekova, Tasnima and Kudinov, Mikhail},
  booktitle={International conference on machine learning},
  pages={8599--8608},
  year={2021},
  organization={PMLR}
}

@article{gal2022image,
  title={An image is worth one word: Personalizing text-to-image generation using textual inversion},
  author={Gal, Rinon and Alaluf, Yuval and Atzmon, Yuval and Patashnik, Or and Bermano, Amit H and Chechik, Gal and Cohen-Or, Daniel},
  journal={arXiv preprint arXiv:2208.01618},
  year={2022}
}

@inproceedings{ruiz2023dreambooth,
  title={Dreambooth: Fine tuning text-to-image diffusion models for subject-driven generation},
  author={Ruiz, Nataniel and Li, Yuanzhen and Jampani, Varun and Pritch, Yael and Rubinstein, Michael and Aberman, Kfir},
  booktitle={Proceedings of the IEEE/CVF conference on computer vision and pattern recognition},
  pages={22500--22510},
  year={2023}
}

@article{song2025omniconsistency,
  title={Omniconsistency: Learning style-agnostic consistency from paired stylization data},
  author={Song, Yiren and Liu, Cheng and Shou, Mike Zheng},
  journal={arXiv preprint arXiv:2505.18445},
  year={2025}
}

@article{yin2025don,
  title={Don't Forget the Nonlinearity: Unlocking Activation Functions in Efficient Fine-Tuning},
  author={Yin, Bo and Yang, Xingyi and Wang, Xinchao},
  journal={arXiv preprint arXiv:2509.13240},
  year={2025}
}

@article{song2025makeanything,
  title={Makeanything: Harnessing diffusion transformers for multi-domain procedural sequence generation},
  author={Song, Yiren and Liu, Cheng and Shou, Mike Zheng},
  journal={arXiv preprint arXiv:2502.01572},
  year={2025}
}

@article{karras2022elucidating,
  title={Elucidating the design space of diffusion-based generative models},
  author={Karras, Tero and Aittala, Miika and Aila, Timo and Laine, Samuli},
  journal={Advances in neural information processing systems},
  volume={35},
  pages={26565--26577},
  year={2022}
}

@article{san2021noise,
  title={Noise estimation for generative diffusion models},
  author={San-Roman, Robin and Nachmani, Eliya and Wolf, Lior},
  journal={arXiv preprint arXiv:2104.02600},
  year={2021}
}

@article{song2023consistency,
  title={Consistency models},
  author={Song, Yang and Dhariwal, Prafulla and Chen, Mark and Sutskever, Ilya},
  year={2023}
}

@article{li2023zero,
  title={Zero-shot medical image translation via frequency-guided diffusion models},
  author={Li, Yunxiang and Shao, Hua-Chieh and Liang, Xiao and Chen, Liyuan and Li, Ruiqi and Jiang, Steve and Wang, Jing and Zhang, You},
  journal={IEEE transactions on medical imaging},
  volume={43},
  number={3},
  pages={980--993},
  year={2023},
  publisher={IEEE}
}

@article{li2025uni,
  title={Uni-LoRA: One Vector is All You Need},
  author={Li, Kaiyang and Han, Shaobo and Su, Qing and Li, Wei and Cai, Zhipeng and Ji, Shihao},
  journal={arXiv preprint arXiv:2506.00799},
  year={2025}
}

@inproceedings{zhu2024task,
  title={Task-customized mixture of adapters for general image fusion},
  author={Zhu, Pengfei and Sun, Yang and Cao, Bing and Hu, Qinghua},
  booktitle={Proceedings of the IEEE/CVF conference on computer vision and pattern recognition},
  pages={7099--7108},
  year={2024}
}

@article{ruderman1993statistics,
  title={Statistics of natural images: Scaling in the woods},
  author={Ruderman, Daniel and Bialek, William},
  journal={Advances in neural information processing systems},
  volume={6},
  year={1993}
}

@article{field1987relations,
  title={Relations between the statistics of natural images and the response properties of cortical cells},
  author={Field, David J},
  journal={Journal of the Optical Society of America A},
  volume={4},
  number={12},
  pages={2379--2394},
  year={1987},
  publisher={OSA}
}

@article{shazeer2017outrageously,
  title={Outrageously large neural networks: The sparsely-gated mixture-of-experts layer},
  author={Shazeer, Noam and Mirhoseini, Azalia and Maziarz, Krzysztof and Davis, Andy and Le, Quoc and Hinton, Geoffrey and Dean, Jeff},
  journal={arXiv preprint arXiv:1701.06538},
  year={2017}
}

@article{lowe2004distinctive,
  title={Distinctive image features from scale-invariant keypoints},
  author={Lowe, David G},
  journal={International journal of computer vision},
  volume={60},
  number={2},
  pages={91--110},
  year={2004},
  publisher={Springer}
}

@article{salimans2022progressive,
  title={Progressive distillation for fast sampling of diffusion models},
  author={Salimans, Tim and Ho, Jonathan},
  journal={arXiv preprint arXiv:2202.00512},
  year={2022}
}

@inproceedings{radford2021learning,
  title={Learning transferable visual models from natural language supervision},
  author={Radford, Alec and Kim, Jong Wook and Hallacy, Chris and Ramesh, Aditya and Goh, Gabriel and Agarwal, Sandhini and Sastry, Girish and Askell, Amanda and Mishkin, Pamela and Clark, Jack and others},
  booktitle={International conference on machine learning},
  pages={8748--8763},
  year={2021},
  organization={PmLR}
}

@inproceedings{houlsby2019parameter,
  title={Parameter-efficient transfer learning for NLP},
  author={Houlsby, Neil and Giurgiu, Andrei and Jastrzebski, Stanislaw and Morrone, Bruna and De Laroussilhe, Quentin and Gesmundo, Andrea and Attariyan, Mona and Gelly, Sylvain},
  booktitle={International conference on machine learning},
  pages={2790--2799},
  year={2019},
  organization={PMLR}
}

@article{guo2020parameter,
  title={Parameter-efficient transfer learning with diff pruning},
  author={Guo, Demi and Rush, Alexander M and Kim, Yoon},
  journal={arXiv preprint arXiv:2012.07463},
  year={2020}
}

@article{fedus2022switch,
  title={Switch transformers: Scaling to trillion parameter models with simple and efficient sparsity},
  author={Fedus, William and Zoph, Barret and Shazeer, Noam},
  journal={Journal of Machine Learning Research},
  volume={23},
  number={120},
  pages={1--39},
  year={2022}
}

@inproceedings{ganjdanesh2024mixture,
  title={Mixture of efficient diffusion experts through automatic interval and sub-network selection},
  author={Ganjdanesh, Alireza and Kang, Yan and Liu, Yuchen and Zhang, Richard and Lin, Zhe and Huang, Heng},
  booktitle={European Conference on Computer Vision},
  pages={54--71},
  year={2024},
  organization={Springer}
}

@article{bai2023qwen,
  title={Qwen technical report},
  author={Bai, Jinze and Bai, Shuai and Chu, Yunfei and Cui, Zeyu and Dang, Kai and Deng, Xiaodong and Fan, Yang and Ge, Wenbin and Han, Yu and Huang, Fei and others},
  journal={arXiv preprint arXiv:2309.16609},
  year={2023}
}

@article{ramesh2022hierarchical,
  title={Hierarchical text-conditional image generation with clip latents},
  author={Ramesh, Aditya and Dhariwal, Prafulla and Nichol, Alex and Chu, Casey and Chen, Mark},
  journal={arXiv preprint arXiv:2204.06125},
  volume={1},
  number={2},
  pages={3},
  year={2022}
}

@article{bengio2013representation,
  title={Representation learning: A review and new perspectives},
  author={Bengio, Yoshua and Courville, Aaron and Vincent, Pascal},
  journal={IEEE transactions on pattern analysis and machine intelligence},
  volume={35},
  number={8},
  pages={1798--1828},
  year={2013},
  publisher={IEEE}
}

@article{kingma2019introduction,
  title={An introduction to variational autoencoders},
  author={Kingma, Diederik P and Welling, Max and others},
  journal={Foundations and Trends{\textregistered} in Machine Learning},
  volume={12},
  number={4},
  pages={307--392},
  year={2019},
  publisher={Now Publishers, Inc.}
}

@article{liu2024routers,
  title={Routers in vision mixture of experts: An empirical study},
  author={Liu, Tianlin and Blondel, Mathieu and Riquelme, Carlos and Puigcerver, Joan},
  journal={arXiv preprint arXiv:2401.15969},
  year={2024}
}

@article{kwakernaak1991modern,
  title={Modern signals and systems},
  author={Kwakernaak, Huibert and Sivan, Raphael},
  journal={NASA STI/Recon Technical Report A},
  volume={91},
  pages={11586},
  year={1991}
}

@article{wan2025wan,
  title={Wan: Open and advanced large-scale video generative models},
  author={Wan, Team and Wang, Ang and Ai, Baole and Wen, Bin and Mao, Chaojie and Xie, Chen-Wei and Chen, Di and Yu, Feiwu and Zhao, Haiming and Yang, Jianxiao and others},
  journal={arXiv preprint arXiv:2503.20314},
  year={2025}
}

@inproceedings{liu2022p,
  title={P-tuning: Prompt tuning can be comparable to fine-tuning across scales and tasks},
  author={Liu, Xiao and Ji, Kaixuan and Fu, Yicheng and Tam, Weng and Du, Zhengxiao and Yang, Zhilin and Tang, Jie},
  booktitle={Proceedings of the 60th Annual Meeting of the Association for Computational Linguistics (Volume 2: Short Papers)},
  pages={61--68},
  year={2022}
}

@article{cao2025task,
  title={Task-Adaptive Parameter-Efficient Fine-Tuning for Weather Foundation Models},
  author={Cao, Shilei and Lin, Hehai and Cheng, Jiashun and Liu, Yang and Li, Guowen and Wang, Xuehe and Zheng, Juepeng and Liang, Haoyuan and Jin, Meng and Qin, Chengwei and others},
  journal={arXiv preprint arXiv:2509.22020},
  year={2025}
}

@article{yu2025dmfft,
  title={DMFFT: improving the generation quality of diffusion models using fast Fourier transform},
  author={Yu, Cuihong and Han, Cheng and Zhang, Chao},
  journal={Scientific Reports},
  volume={15},
  number={1},
  pages={10200},
  year={2025},
  publisher={Nature Publishing Group UK London}
}

@article{tivnan2025fourier,
  title={Fourier diffusion models: A method to control mtf and nps in score-based stochastic image generation},
  author={Tivnan, Matthew and Teneggi, Jacopo and Lee, Tzu-Cheng and Zhang, Ruoqiao and Boedeker, Kirsten and Cai, Liang and Gang, Grace J and Sulam, Jeremias and Stayman, J Webster},
  journal={IEEE transactions on medical imaging},
  year={2025},
  publisher={IEEE}
}

@inproceedings{valadarsky2017learning,
  title={Learning to route},
  author={Valadarsky, Asaf and Schapira, Michael and Shahaf, Dafna and Tamar, Aviv},
  booktitle={Proceedings of the 16th ACM workshop on hot topics in networks},
  pages={185--191},
  year={2017}
}

@inproceedings{chen2024find,
  title={Find: Fine-tuning initial noise distribution with policy optimization for diffusion models},
  author={Chen, Changgu and Yang, Libing and Yang, Xiaoyan and Chen, Lianggangxu and He, Gaoqi and Wang, Changbo and Li, Yang},
  booktitle={Proceedings of the 32nd ACM International Conference on Multimedia},
  pages={6735--6744},
  year={2024}
}

@article{arora2024low,
  title={Low-frequency adaptation-deep neural network-based domain adaptation approach for shaft imbalance fault diagnosis},
  author={Arora, Jatin Kumar and Rajagopalan, Sudhar and Singh, Jaskaran and Purohit, Ashish},
  journal={Journal of Vibration Engineering \& Technologies},
  volume={12},
  number={1},
  pages={375--394},
  year={2024},
  publisher={Springer}
}

@inproceedings{xu2020learning,
  title={Learning in the frequency domain},
  author={Xu, Kai and Qin, Minghai and Sun, Fei and Wang, Yuhao and Chen, Yen-Kuang and Ren, Fengbo},
  booktitle={Proceedings of the IEEE/CVF conference on computer vision and pattern recognition},
  pages={1740--1749},
  year={2020}
}

@inproceedings{hu2020face,
  title={Face super-resolution guided by 3d facial priors},
  author={Hu, Xiaobin and Ren, Wenqi and LaMaster, John and Cao, Xiaochun and Li, Xiaoming and Li, Zechao and Menze, Bjoern and Liu, Wei},
  booktitle={European Conference on Computer Vision},
  pages={763--780},
  year={2020},
  organization={Springer}
}

@inproceedings{ji2025sonic,
  title={Sonic: Shifting focus to global audio perception in portrait animation},
  author={Ji, Xiaozhong and Hu, Xiaobin and Xu, Zhihong and Zhu, Junwei and Lin, Chuming and He, Qingdong and Zhang, Jiangning and Luo, Donghao and Chen, Yi and Lin, Qin and others},
  booktitle={Proceedings of the Computer Vision and Pattern Recognition Conference},
  pages={193--203},
  year={2025}
}

@article{ji2024realtalk,
  title={Realtalk: Real-time and realistic audio-driven face generation with 3d facial prior-guided identity alignment network},
  author={Ji, Xiaozhong and Lin, Chuming and Ding, Zhonggan and Tai, Ying and Zhu, Junwei and Hu, Xiaobin and Luo, Donghao and Ge, Yanhao and Wang, Chengjie},
  journal={arXiv preprint arXiv:2406.18284},
  year={2024}
}

@article{xiaobin2025vtbench,
  title={VTBench: Comprehensive Benchmark Suite Towards Real-World Virtual Try-on Models},
  author={Xiaobin, Hu and Yujie, Liang and Donghao, Luo and Xu, Peng and Jiangning, Zhang and Junwei, Zhu and Chengjie, Wang and Yanwei, Fu},
  journal={arXiv preprint arXiv:2505.19571},
  year={2025}
}

@inproceedings{lee2024multi,
  title={Multi-architecture multi-expert diffusion models},
  author={Lee, Yunsung and Kim, JinYoung and Go, Hyojun and Jeong, Myeongho and Oh, Shinhyeok and Choi, Seungtaek},
  booktitle={Proceedings of the AAAI Conference on Artificial Intelligence},
  volume={38},
  number={12},
  pages={13427--13436},
  year={2024}
}

@inproceedings{park2024switch,
  title={Switch diffusion transformer: Synergizing denoising tasks with sparse mixture-of-experts},
  author={Park, Byeongjun and Go, Hyojun and Kim, Jin-Young and Woo, Sangmin and Ham, Seokil and Kim, Changick},
  booktitle={European Conference on Computer Vision},
  pages={461--477},
  year={2024},
  organization={Springer}
}

@article{fei2024scaling,
  title={Scaling diffusion transformers to 16 billion parameters},
  author={Fei, Zhengcong and Fan, Mingyuan and Yu, Changqian and Li, Debang and Huang, Junshi},
  journal={arXiv preprint arXiv:2407.11633},
  year={2024}
}

@article{park2023denoising,
  title={Denoising task routing for diffusion models},
  author={Park, Byeongjun and Woo, Sangmin and Go, Hyojun and Kim, Jin-Young and Kim, Changick},
  journal={arXiv preprint arXiv:2310.07138},
  year={2023}
}

@inproceedings{yang2023diffusion,
  title={Diffusion probabilistic model made slim},
  author={Yang, Xingyi and Zhou, Daquan and Feng, Jiashi and Wang, Xinchao},
  booktitle={Proceedings of the IEEE/CVF Conference on computer vision and pattern recognition},
  pages={22552--22562},
  year={2023}
}
\bibliographystyle{icml2026}

\newpage
\appendix
\onecolumn
\section{Frequency-Energy Indicator Is Not a Proxy for Timestep}
\label{sec:fei_analysis}

\begin{figure}[htbp]
    \centering
    \includegraphics[width=0.75\linewidth]{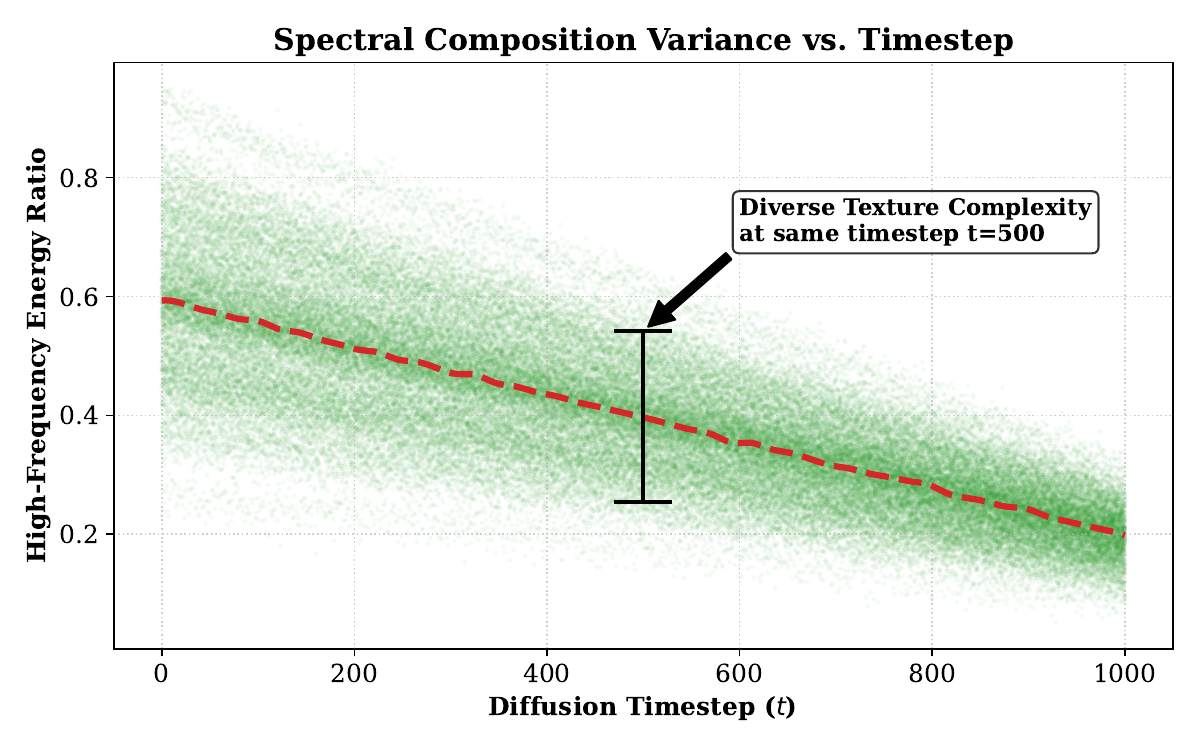}
    \caption{Instance-wise spectral variance.}
    \label{fig:fei_variance}
\end{figure}

To validate the design motivation of FeRA, we investigate whether the Frequency-Energy Indicator (FEI) merely acts as a proxy for the diffusion timestep $t$. As shown in Fig.~\ref{fig:fei_variance}, we visualized the spectral evolution of 100 generation trajectories. 
\textbf{Consistent with standard diffusion notation where $t=0$ represents the clean image and $t=T$ represents pure noise, the population mean exhibits a monotonic decrease, which means opposite direction to Fig.~\ref{fig:freq-evo}} This trend confirms the expected physical behavior: structural information degrades as noise intensity increases. 
However, this mean trend represents the limit of what static time embeddings can capture a global, uniform routing policy shared across all samples.

Crucially, the scatter plot reveals significant instance-specific variance around this mean. As annotated at $t=500$, the high-frequency ratio spans a wide range (approx. $0.25$ to $0.60$), clearly distinguishing texture-rich instances from structurally simple ones at the exact same denoising stage. This empirical evidence confirms that FEI is not a proxy for time; unlike time-based methods, FeRA leverages this content-dependent variance to dynamically allocate expert capacity based on the actual complexity of the instance.

\section{Theoretical Analysis of Frequency-Energy Routing}
\label{sec:theory_formulation}

In this section, we provide a formal theoretical comparison between standard Time-Dependent Mixture-of-Experts (Time-MoE) and the proposed Frequency-Energy Routing (FeRA). Both of them use soft routing. We prove that FeRA implements a state-dependent control policy that strictly generalizes time-dependent policies, offering a superior optimization landscape by filtering updates on noise-dominated frequency bands.

\subsection{Routing Manifolds: Open-Loop vs. Closed-Loop}

Let $\mathcal{Z} \subseteq \mathbb{R}^{d}$ denote the latent space of the diffusion model, and let $\{x_t\}_{t=0}^T$ represent the stochastic trajectory of the diffusion process.

\textbf{Definition 1 (Time-Dependent Routing).} A standard Time-MoE defines the routing weights $\mathbf{w} \in \Delta^{M-1}$ (where $\Delta$ is the simplex) as a function of the scalar timestep $t$:
\begin{equation}
    \Phi_{time}: [0, T] \to \Delta^{M-1}, \quad \mathbf{w}_t = \text{Softmax}(\mathbf{W}_{\tau} \cdot \psi(t))
\end{equation}
where $\psi(t)$ is a fixed temporal embedding. This represents an \textit{open-loop} control system where the expert selection is independent of the instantaneous state $x_t$. The decision boundary for Expert $k$ is defined by the level set $\{t \mid (\mathbf{w}_t)_k > \gamma\}$, which corresponds to fixed hyperplanes orthogonal to the time axis.

\textbf{Definition 2 (Frequency-Energy Routing).} FeRA defines the routing weights as a function of the spectral energy state of the latent $x_t$:
\begin{equation}
    \Phi_{FeRA}: \mathcal{Z} \to \Delta^{M-1}, \quad \mathbf{w}_{freq} = \text{Softmax}\left(\frac{g_\phi(\mathcal{F}(x_t))}{\tau}\right)
\end{equation}
where $\mathcal{F}: \mathcal{Z} \to \mathbb{R}^n$ is the Frequency-Energy Indicator (FEI) operator extracting band-wise energy, and $g_\phi$ is a learnable projection. This represents a \textit{closed-loop} feedback system.

\textbf{Proposition 1 (Generalization Capability).} 
Since the expected spectral energy $\mathbb{E}[\mathcal{F}(x_t)]$ is monotonic with respect to $t$ (due to the diffusion schedule $\alpha_t$), there exists a mapping $h$ such that $\mathbb{E}[\Phi_{FeRA}(x_t)] \approx \Phi_{time}(t)$. However, since the instantaneous energy $\mathcal{F}(x_t)$ contains variance not captured by $t$ (i.e., entropy $H(\mathcal{F}(x_t) | t) > 0$), $\Phi_{FeRA}$ operates on a strictly richer manifold. $\Phi_{time}$ is a degenerate case of $\Phi_{FeRA}$ where the input is replaced by its population expectation.

\subsection{Gradient Orthogonality and Noise Suppression}

We analyze the optimization behavior by examining the gradient applied to a specific expert adapter $\theta_k$ assigned to a high-frequency band. Consider the standard denoising objective $\mathcal{L} = \mathbb{E}_{x_0, \epsilon, t} [ \| \epsilon - \epsilon_\theta(x_t, t) \|^2 ]$.

Let the residual error be decomposed into frequency components. The gradient update for expert $k$ is proportional to the router activation $w^{(k)}$:
\begin{equation}
    \nabla_{\theta_k} \mathcal{L} \propto w^{(k)} \cdot \frac{\partial \mathcal{L}}{\partial \text{Output}}
\end{equation}

\textbf{Theorem 1 (Gradient Noise in Time-MoE).} 
In Time-MoE, let $T_k$ be the time interval where expert $k$ is active (i.e., $w^{(k)}(t) \approx 1$). For a specific sample $x_t$, if the actual signal energy in band $k$ is zero (denoted as $S_k(x_t) = 0$), the neural network attempts to fit the pure noise component $\epsilon_k$.
Since $w^{(k)}_{time}(t)$ is determined solely by $t$, if $t \in T_k$, then $w^{(k)}_{time} \gg 0$. Consequently:
\begin{equation}
    \mathbb{E} [ \| \nabla_{\theta_k} \mathcal{L} \| \mid S_k(x_t) = 0, t \in T_k ] > 0
\end{equation}
This implies the expert $\theta_k$ receives non-zero gradient updates to fit pure Gaussian noise, leading to overfitting and optimization instability.

\textbf{Theorem 2 (Sparsity in FeRA).} 
In FeRA, the activation $w^{(k)}_{FeRA}$ is conditioned on the FEI, which approximates the signal-to-noise ratio. If a band $k$ lacks energy (i.e., $S_k(x_t) \approx 0$), the FEI vector reflects this sparsity, causing the router to suppress expert $k$:
\begin{equation}
    S_k(x_t) \to 0 \implies \mathcal{F}(x_t)_k \to 0 \implies w^{(k)}_{FeRA} \to 0
\end{equation}
Therefore, the gradient magnitude is bounded by the signal presence:
\begin{equation}
    \lim_{S_k \to 0} \| \nabla_{\theta_k} \mathcal{L} \|_{FeRA} = 0
\end{equation}
This property, which we term \textit{Signal-Gradient Orthogonality}, ensures that FeRA experts are only updated when valid structural information is present, effectively performing dynamic curriculum learning unavailable to time-dependent routers.

\section{Generalization to Non-Natural Domains}
\label{sec:non_natural_generalization}

A valid concern regarding frequency-aware methods is whether they rely excessively on natural image statistics (e.g., the $1/f$ power law), potentially limiting generalization to non-natural domains such as line art, cel-shaded anime, or abstract textures. We clarify that FeRA is designed to be domain-agnostic by utilizing the Frequency-Energy Indicator (FEI) as a sensing mechanism rather than a hard constraint. The FEI acts as a descriptive state indicator that measures the actual spectral composition of the latent features, regardless of whether that composition adheres to natural statistics.

\begin{figure}[t]
    \centering
    \includegraphics[width=\linewidth]{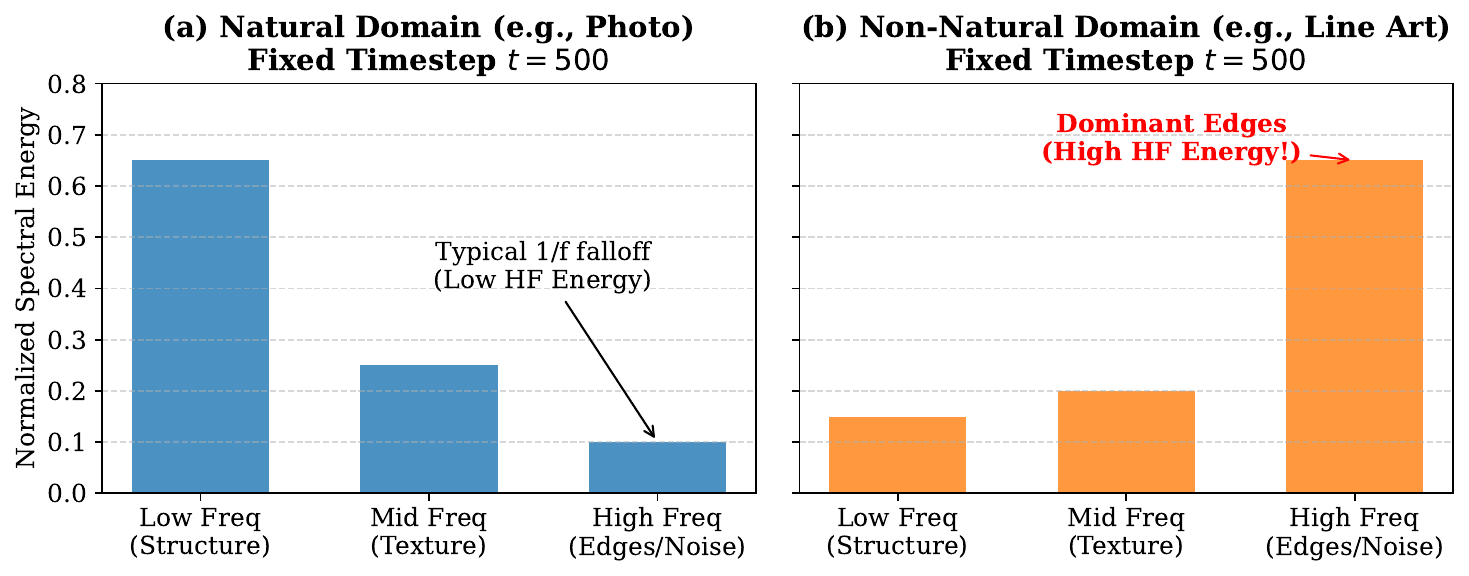}
    \caption{Spectral signature contrast at a fixed timestep ($t=500$).}
    \label{fig:domain_generalization}
\end{figure}

The key to FeRA's robustness lies in its learnable routing network. During fine-tuning on a specific target domain, the router parameters are optimized to map the observed spectral signatures to the most effective expert combination. For instance, as illustrated in Fig.~\ref{fig:domain_generalization}, domains like line art are inherently dominated by high-frequency edge information even at intermediate timesteps (e.g., $t=500$), contrasting sharply with the low-frequency dominance of natural images. The router learns to interpret these high FEI values as essential structural features rather than noise, adaptively activating high-frequency experts. This contrasts with fixed priors that might suppress such signals.

Consequently, rather than restricting generalization, frequency-awareness enhances the model's adaptability to out-of-distribution domains. While static, time-based baselines apply a uniform denoising policy blind to the domain's spectral shift, FeRA's instance-aware routing allows it to adjust its computational focus based on the specific complexity of the target domain, ensuring effective adaptation across both natural and artistic distributions.

\section{Experiment Setting}
\subsection{MLLM-Judge Prompt}
In experiment we evaluate stylistic fidelity(Style Score) using an
MLLM-based style assessor.
The model is prompted to judge the stylistic attributes of the generated images.For transparency and reproducibility, we include the exact prompt used in all evaluations
below.
\begin{tcolorbox}[
    colback=gray!3,
    colframe=black!70,
    boxrule=0.55pt,
    arc=0.8mm,
    left=2mm,right=2mm,top=1.6mm,bottom=1.6mm
]
\ttfamily\small
You are a strict T2I style judge.\\
Evaluate ONLY the STYLE qualities of the output image.\\
Use style cues from the PROMPT (and optional STYLE\_TEXT / STYLE refs).\\
Ignore all semantic correctness.\\[0.45em]

Return a compact JSON with: style\_faithfulness, style\_intensity,\\
palette\_match, lighting\_mood, texture\_pattern, artifacts, overall\_style.\\
Each score is in [0,10] (integer or one decimal). No extra text.\\[0.55em]

Definitions:\\
- style\_faithfulness: reflection of style cues (medium, era, motifs).\\
- style\_intensity: stylization strength (not too weak or excessive).\\
- palette\_match: color palette and tone mapping.\\
- lighting\_mood: lighting style and ambience.\\
- texture\_pattern: local texture/strokes/pattern effects.\\
- artifacts: fewer visual artifacts $\Rightarrow$ higher score.\\
- overall\_style = 0.30*style\_faithfulness + 0.15*style\_intensity\\
\ \ \ \ \ \ \ \ \ + 0.20*palette\_match + 0.15*lighting\_mood\\
\ \ \ \ \ \ \ \ \ + 0.20*texture\_pattern.
\end{tcolorbox}

\subsection{Training Setting}
For clarity and reproducibility, we summarize the common training configuration used across all experiments in Table~\ref{tab:exp-setting}.

\begin{table*}[htbp]
\centering
\small
\renewcommand{\arraystretch}{1.22}
\setlength{\tabcolsep}{6pt}
\caption{Common experiment settings used throughout our study.}
\begin{tabular}{p{3.2cm} p{10.8cm}}
\hline
\textbf{Category} & \textbf{Setting} \\
\hline

Base Model & frozen VAE and text encoder \\

LoRA Config & applied to attention modules (\texttt{to\_q}, \texttt{to\_k}, \texttt{to\_v}, \texttt{to\_out}) \\

Resolution & $512 \times 512$ (resize + center/random crop) \\

Batch Size & 16 per device \\

Training Steps & 5000 \\

Optimizer & AdamW, LR = $1{\times}10^{-4}$, weight decay = 0.01 \\

Warmup Steps & 500 \\

Scheduler & DDPM scheduler with standard noise schedule \\

Mixed Precision & fp16 \\

Grad Norm Clip & 1.0 \\

Inference & 30-step DDPM sampling\\

Hardware & NVIDIA H100 GPUs \\
\hline
\end{tabular}

\label{tab:exp-setting}
\end{table*}

\section{Other Experiment Result}
\subsection{Text-to-Image Style Adaptation}
To further examine the generality of our training pipeline, we extend the
text-to-image style adaptation experiments to multiple diffusion backbones
with distinct latent resolutions, denoising trajectories, and text–image
alignment capabilities. This broader evaluation helps disentangle whether
the observed performance gains truly stem from our frequency–energy–guided
fine-tuning strategy rather than from accidental synergy with a particular
pretrained model. By comparing models that differ substantially in their
architectural design and training dynamics, we can more reliably assess the
stability and scalability of our method.

As summarized in Tab.~\ref{tab:sd15_sdxl_compact}, the performance trends
remain consistent across Stable Diffusion~1.5 and Stable Diffusion~XL.
Despite the large gap in model capacity and latent-space structure, our
approach produces clear improvements in perceptual quality, stylistic
faithfulness, and controllability. At the same time, semantic alignment
remains competitive, suggesting that the additional style expressiveness
does not come at the cost of prompt consistency. These results indicate that
our fine-tuning strategy transfers reliably across diffusion families and
retains its benefits even as the backbone scale or architecture changes.

\begin{figure*}[!t]
    \centering
    \includegraphics[width=\linewidth]{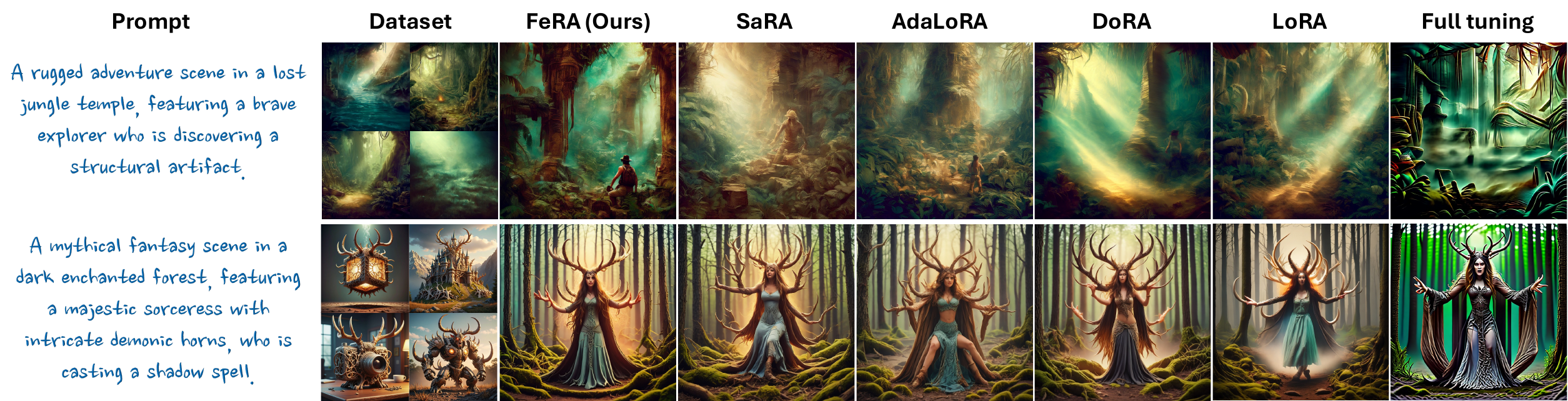}
    \caption{Comparison of the generated images between different PEFT methods in other datasets.}
    \label{fig:exp_visuala}
\end{figure*}

\begin{figure*}[!t]
    \centering
    \includegraphics[width=\linewidth]{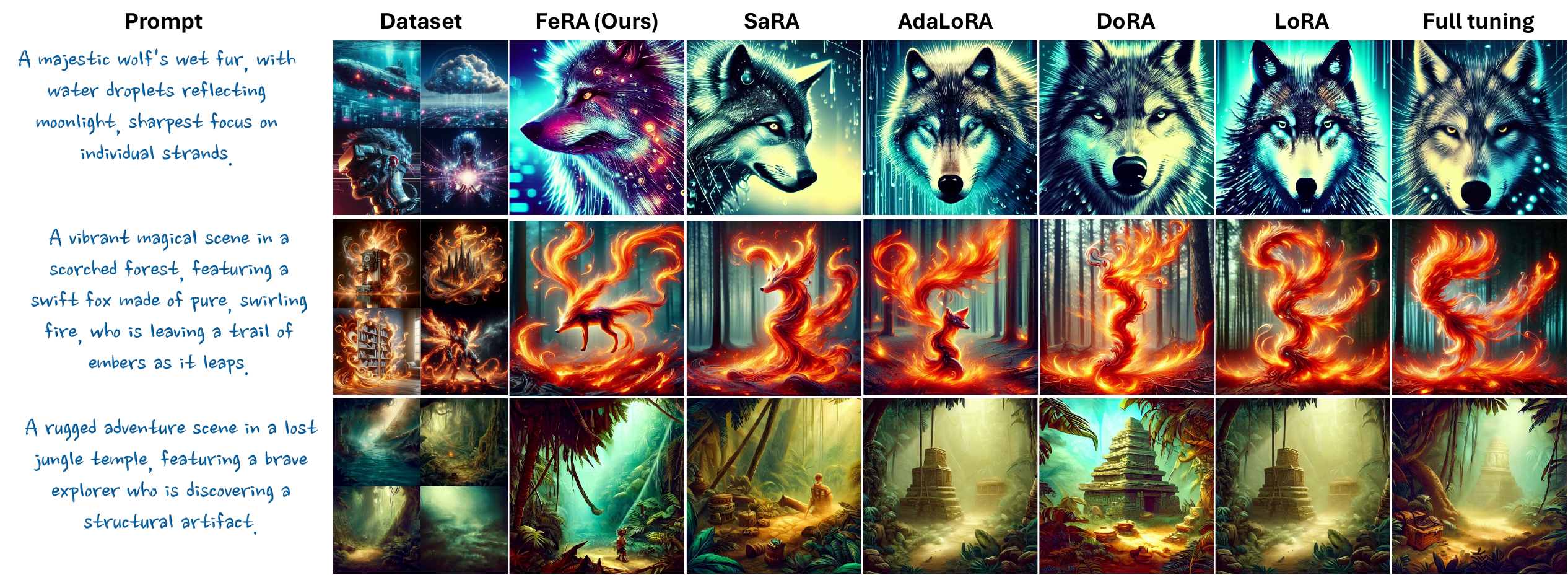}
    \caption{Comparison of the generated images between different PEFT methods(SD1.5).}
    \label{fig:exp_visuala15}
\end{figure*}

\begin{figure*}[!t]
    \centering
    \includegraphics[width=\linewidth]{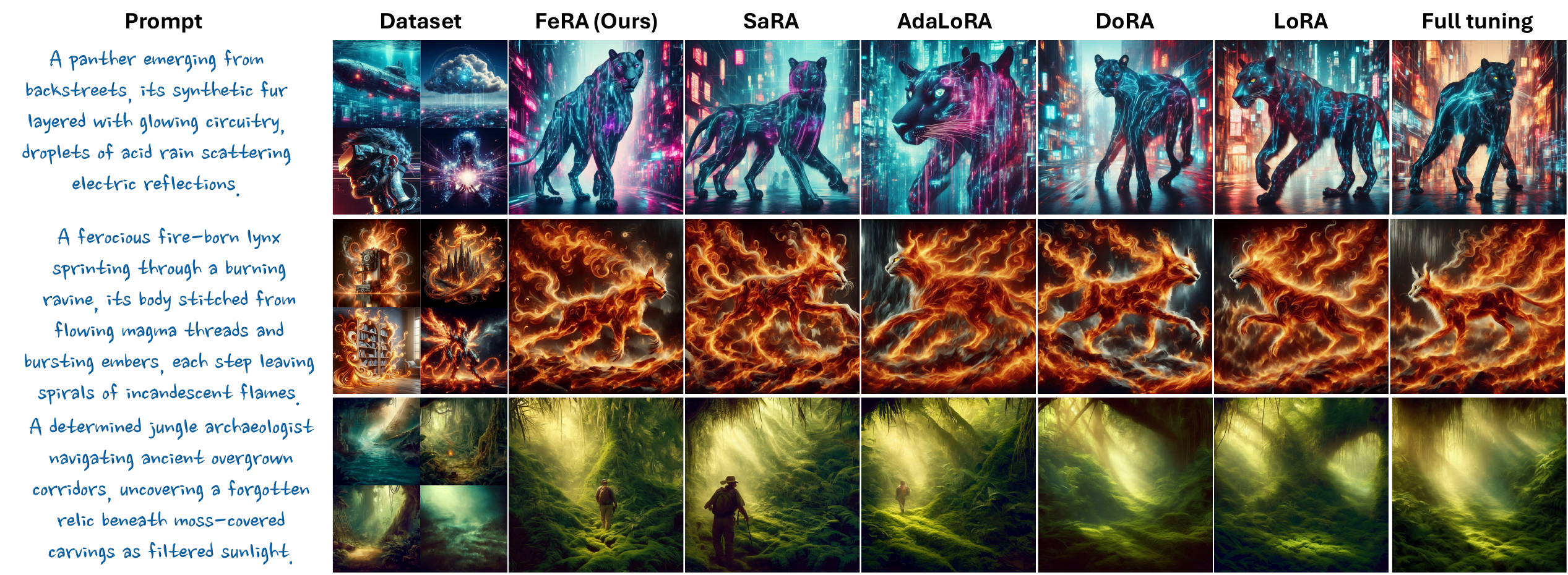}
    \caption{Comparison of the generated images between different PEFT methods(SD3.0).}
    \label{fig:exp_visuala3}
\end{figure*}

And we also show others qualitative result in other datasets or backbones in Fig.~\ref{fig:exp_visuala}, \ref{fig:exp_visuala15} and \ref{fig:exp_visuala3}.

\begin{table*}[htbp]
\small
\renewcommand{\arraystretch}{0.93}
\setlength{\tabcolsep}{2.3pt}
\centering
\caption{Comparison with different parameter-efficient fine-tuning methods on Stable Diffusion 1.5 and SDXL. \best{Orange bold} = best \second{Light orange italic} = second best. \textbf{\textit{Notice:} the training parameter is same for fair comparison by controlling the rank of FeRA.}}
\vspace{-2mm}
\resizebox{\linewidth}{!}{
\begin{tabular}{c|c|c|ccc|ccc|ccc|ccc|ccc}
\toprule
\multirow{2}{*}{Backbone} & \multirow{2}{*}{Params} & \multirow{2}{*}{Method}
& \multicolumn{3}{c}{Barbie}
& \multicolumn{3}{c}{Cyberpunk}
& \multicolumn{3}{c}{Expedition}
& \multicolumn{3}{c}{Hornify}
& \multicolumn{3}{c}{Elementfire} \\
& & & CLIP $\uparrow$ & FID $\downarrow$ & Style $\uparrow$
& CLIP $\uparrow$ & FID $\downarrow$ & Style $\uparrow$
& CLIP $\uparrow$ & FID $\downarrow$ & Style $\uparrow$
& CLIP $\uparrow$ & FID $\downarrow$ & Style $\uparrow$
& CLIP $\uparrow$ & FID $\downarrow$ & Style $\uparrow$ \\
\midrule

\multirow{16}{*}{SD 1.5}
& \multirow{5}{*}{5M}
& LoRA & \best{35.19} & 194.73 & 8.37 & 33.16 & \second{129.63} & 8.14 & 31.51 & 128.73 & 8.46 & 31.08 & 163.02 & 8.48 & \second{31.59} & 156.95 & 8.47 \\
& & DoRA & 35.12 & 193.90 & 8.36 & \best{33.20} & 131.51 & \second{8.19} & 31.53 & 129.13 & 8.47 & 31.15 & 164.24 & 8.50 & \best{31.62} & 158.59 & 8.49 \\
& & AdaLoRA & \second{35.12} & 194.30 & 8.38 & 33.12 & 131.51 & 8.17 & 31.53 & 129.00 & \second{8.50} & 31.15 & \second{162.81} & 8.49 & 31.55 & \second{156.62} & 8.51 \\
& & SaRA & 35.01 & \second{190.81} & \second{8.39} & \second{33.18} & 138.01 & 8.18 & \second{31.56} & \second{128.50} & 8.49 & \second{31.18} & 163.50 & \second{8.51} & 31.55 & 166.00 & \second{8.52} \\
& &  \textbf{FeRA (Ours)} & 34.34 & \best{184.98} & \best{8.43} & 32.96 & \best{126.40} & \best{8.21} & \best{31.66} & \best{127.03} & \best{8.53} & \best{31.20} & \best{162.28} & \best{8.54} & 31.12 & \best{154.92} & \best{8.55} \\
\cmidrule(lr){2-18}
& \multirow{5}{*}{20M}
& LoRA & \best{35.31} & 181.15 & 8.58 & 33.25 & 122.32 & \second{8.04} & 31.60 & 120.10 & 8.05 & 31.19 & 151.75 & 7.94 & \best{31.70} & 146.50 & 7.72 \\
& & DoRA & 35.25 & 180.32 & 8.01 & 33.22 & 122.32 & 7.59 & 31.63 & 120.12 & \second{8.13} & 31.25 & 152.12 & \second{8.34} & 31.64 & 155.32 & \second{8.27} \\
& & AdaLoRA & \second{35.25} & 180.70 & \second{8.60} & 33.22 & \second{122.30} & 7.83 & 31.63 & 119.95 & 7.58 & 31.25 & \second{151.45} & 8.31 & 31.62 & \second{146.11} & 8.17 \\
& & SaRA & 35.15 & \second{177.50} & 8.06 & \second{33.28} & 127.80 & 7.70 & \second{31.67} & \second{119.50} & 7.96 & \second{31.29} & 152.18 & 7.90 & \second{31.66} & 155.09 & 7.61 \\
& &  \textbf{FeRA (Ours)} & 34.50 & \best{172.03} & \best{8.74} & \best{33.38} & \best{117.90} & \best{8.32} & \best{31.78} & \best{118.11} & \best{8.55} & \best{31.32} & \best{150.91} & \best{8.71} & 31.25 & \best{143.62} & \best{8.60} \\
\cmidrule(lr){2-18}
& \multirow{5}{*}{50M}
& LoRA & \best{35.35} & 183.42 & 8.26 & \second{33.28} & 123.55 & 7.58 & 31.62 & 119.21 & 8.22 & 31.22 & 150.89 & \second{8.62} & \best{31.73} & \second{145.58} & 8.09 \\
& & DoRA & \second{35.29} & 182.63 & \second{8.63} & 33.25 & 123.52 & 7.43 & 31.65 & 119.22 & 8.44 & 31.28 & 151.63 & 7.69 & \second{31.69} & 154.72 & \second{8.21} \\
& & AdaLoRA & 35.29 & 182.92 & 8.16 & 33.25 & \second{123.35} & \second{8.19} & 31.65 & 119.05 & \second{8.52} & 31.28 & \second{150.47} & 8.24 & 31.69 & \best{145.54} & 7.83 \\
& & SaRA & 35.19 & \second{179.72} & 8.27 & \best{33.31} & 129.14 & 7.59 & \second{31.69} & \second{118.61} & \second{8.53} & \second{31.32} & 151.83 & 8.21 & 31.49 & 154.32 & 7.57 \\
& &  \textbf{FeRA (Ours)} & 34.55 & \best{174.21} & \best{8.76} & 33.12 & \best{119.13} & \best{8.34} & \best{31.80} & \best{117.21} & \best{8.62} & \best{32.73} & \best{150.13} & \best{8.66} & 31.29 & 152.48 & \best{8.54} \\
\cmidrule(lr){2-18}
& \multirow{1}{*}{860M}
& Full-Tuning
& 34.48 & 176.50 & 8.70
& 33.05 & 121.20 & 8.30
& 31.75 & 119.00 & 8.58
& 31.30 & 151.80 & 8.63
& 31.20 & 154.00 & 8.51 \\
\midrule

\multirow{16}{*}{SDXL}
& \multirow{5}{*}{5M}
& LoRA & \second{35.88} & 188.41 & 8.53 & 33.45 & 157.92 & 8.38 & 31.94 & 141.52 & 8.60 & 32.45 & 155.82 & 8.66 & 32.32 & 173.73 & 8.67 \\
& & DoRA & 35.83 & 186.78 & 8.52 & 33.48 & 156.21 & 8.41 & 31.96 & 140.28 & 8.63 & 32.48 & 153.94 & 8.65 & 32.35 & 171.35 & 8.68 \\
& & AdaLoRA & 35.85 & 187.60 & \second{8.55} & 33.49 & 157.12 & 8.40 & 31.97 & 140.92 & 8.62 & 32.48 & 154.90 & 8.66 & 32.32 & 172.55 & \second{8.70} \\
& & SaRA & 35.76 & \second{183.25} & 8.54 & \second{33.51} & \second{152.69} & \second{8.43} & \second{31.98} & \second{138.15} & \second{8.64} & \best{32.51} & \second{151.07} & \second{8.67} & \second{32.38} & \second{168.22} & 8.69 \\
& &  \textbf{FeRA (Ours)} & \best{37.86} & \best{178.67} & \best{8.57} & \best{33.61} & \best{148.88} & \best{8.44} & \best{32.08} & \best{134.69} & \best{8.67} & \second{32.49} & \best{147.29} & \best{8.70} & \best{32.48} & \best{163.51} & \best{8.72} \\
\cmidrule(lr){2-18}

& \multirow{5}{*}{20M}
& LoRA & \second{35.99} & 175.23 & 8.36 & 33.57 & 146.84 & 8.74 & 32.07 & 131.42 & 8.44 & 32.58 & 144.90 & 8.28 & 32.45 & 161.42 & 8.05 \\
& & DoRA & 35.94 & 173.73 & 8.87 & 33.60 & 145.72 & 7.97 & 32.09 & 130.17 & 7.88 & 32.61 & 143.12 & \second{8.53} & 32.45 & 159.15 & \second{8.93} \\
& & AdaLoRA & 35.94 & 174.25 & 8.46 & 33.60 & 146.45 & \second{8.77} & 32.09 & 130.82 & \second{8.55} & 32.61 & 144.00 & 8.66 & 32.45 & 160.22 & 8.64 \\
& & SaRA & 35.87 & \second{170.44} & \second{8.88} & \second{33.63} & \second{141.93} & 7.91 & \second{32.11} & \best{128.26} & 8.22 & \best{32.64} & \second{140.23} & 8.30 & \second{32.51} & \second{156.25} & 8.27 \\
& &  \textbf{FeRA (Ours)} & \best{36.97} & \best{166.61} & \best{8.94} & \best{34.73} & \best{138.45} & \best{8.82} & \best{32.21} & \second{128.29} & \best{8.68} & \second{32.62} & \best{136.38} & \best{8.92} & \best{33.70} & \best{152.11} & \best{8.95} \\
\cmidrule(lr){2-18}

& \multirow{5}{*}{50M}
& LoRA & \best{36.01} & 176.41 & \second{9.09} & 33.59 & 147.19 & 8.45 & 32.08 & 130.18 & 8.23 & 32.59 & 144.24 & \second{9.11} & 32.46 & 160.58 & 8.23 \\
& & DoRA & 35.96 & 174.96 & 8.42 & 33.62 & 146.13 & 8.24 & 32.10 & 129.54 & 8.03 & 32.62 & 142.48 & 9.05 & 32.46 & 158.52 & 8.49 \\
& & AdaLoRA & 35.96 & 175.57 & 8.63 & 33.62 & 147.51 & \second{8.82} & 32.10 & 130.22 & 8.80 & 32.62 & 143.35 & 9.03 & 32.46 & 159.26 & \second{8.85} \\
& & SaRA & 35.89 & \second{171.66} & 8.89 & \second{33.65} & \second{143.52} & 8.57 & \second{32.12} & \second{127.66} & \second{8.24} & \second{32.65} & \second{139.61} & 8.98 & \second{32.52} & \second{155.56} & 8.72 \\
& &  \textbf{FeRA (Ours)} & \second{35.99} & \best{167.33} & \best{9.12} & \best{33.75} & \best{139.51} & \best{9.05} & \best{33.22} & \best{124.21} & \best{8.99} & \best{33.63} & \best{136.11} & \best{9.23} & \best{33.62} & \best{151.46} & \best{8.86} \\
\cmidrule(lr){2-18}

& \multirow{1}{*}{3.5B}
& Full-Tuning
& 36.00 & 170.80 & 9.10
& 33.72 & 143.00 & 8.80
& 32.15 & 126.50 & 8.90
& 32.65 & 139.00 & 9.00
& 32.55 & 153.00 & 8.85 \\
\bottomrule

\end{tabular}}
\vspace{-0.35em}
\label{tab:sd15_sdxl_compact}
\vspace{-0.6em}
\end{table*}

\subsection{Image Customization}
We further study instance-level image customization under DreamBooth fine-tuning on two backbones, Stable Diffusion~1.5 and Stable Diffusion~3.0. This setting evaluates how well different PEFT approaches preserve the identity of a specific object while retaining prompt alignment. Tab.~\ref{tab:clip_results_sd15} and
\ref{tab:clip_results_sd3} report CLIP-based image alignment (CLIP-I) and text alignment (CLIP-T).

Across both backbones, FeRA consistently achieves the best or second-best instance faithfulness (CLIP-I) while maintaining competitive or superior text alignment (CLIP-T) compared to LoRA, DoRA, AdaLoRA and SaRA. The trend is stable on both the lighter SD~1.5 and the stronger SD~3.0 models, indicating that our framework provides robust benefits for DreamBooth-style image customization across different backbone capacities.

\begin{figure*}[!t]
    \centering
    \includegraphics[width=\linewidth]{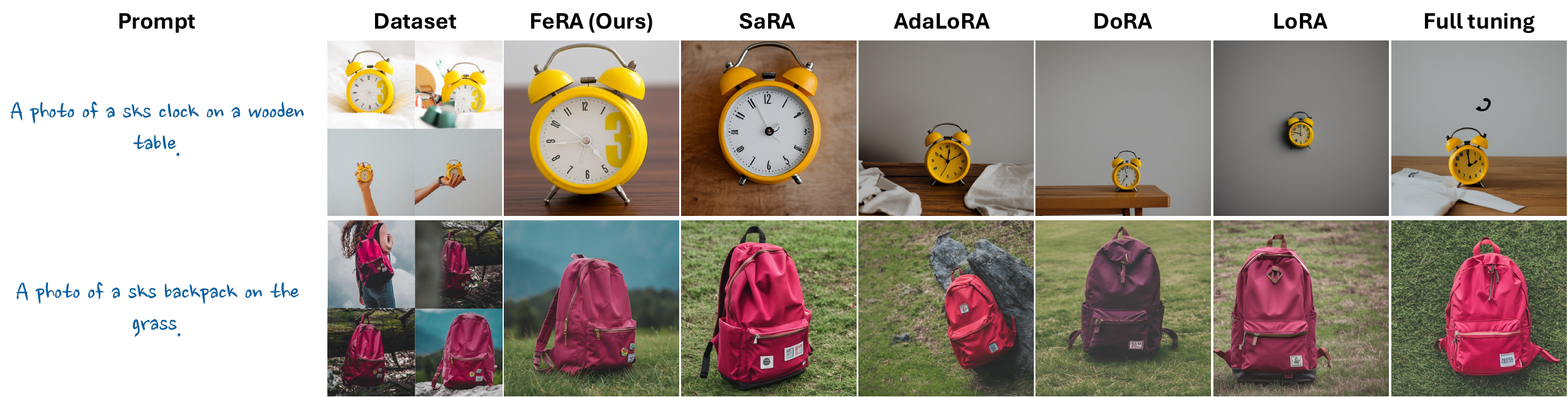}
    \caption{Comparison of the image customization between different PEFT methods.}
    \label{fig:dba}
\end{figure*}

And we also show others qualitative result in other datasets in Fig.~\ref{fig:dba}.

\begin{table*}[t]
\centering
\caption{
Quantitative comparison between different PEFT methods on image customization (Stable Diffusion 1.5).
\textbf{\textcolor{topone}{Red bold}} = best, 
\textit{\textcolor{toptwo}{orange italic}} = second best.
}
\vspace{-2mm}
\small
\setlength{\tabcolsep}{4pt}
\renewcommand{\arraystretch}{1.12}
\resizebox{\linewidth}{!}{%
\begin{tabular}{l|cc|cc|cc|cc|cc}
\toprule
\multirow{2}{*}{Methods} &
\multicolumn{2}{c|}{Dog} &
\multicolumn{2}{c|}{Clock} &
\multicolumn{2}{c|}{Backpack} &
\multicolumn{2}{c|}{Toy Duck} &
\multicolumn{2}{c}{Teapot} \\
\cmidrule(lr){2-11}
 & CLIP-I~$\uparrow$ & CLIP-T~$\uparrow$
 & CLIP-I~$\uparrow$ & CLIP-T~$\uparrow$
 & CLIP-I~$\uparrow$ & CLIP-T~$\uparrow$
 & CLIP-I~$\uparrow$ & CLIP-T~$\uparrow$
 & CLIP-I~$\uparrow$ & CLIP-T~$\uparrow$ \\
\midrule
Dreambooth + Full-tuning & 0.787 & 24.09 & 0.788 & 23.11 & 0.653 & 24.15 & 0.787 & 23.93 & 0.752 & 24.21 \\
Dreambooth + LoRA        & 0.894 & 23.76 & 0.912 & 21.58 & \textit{\textcolor{toptwo}{0.918}} & 25.12 & 0.906 & 23.74 & 0.909 & 23.54 \\
Dreambooth + DoRA        & \textit{\textcolor{toptwo}{0.899}} & 23.59 & \textit{\textcolor{toptwo}{0.914}} & 21.82 & 0.911 & \textit{\textcolor{toptwo}{25.32}} & \textit{\textcolor{toptwo}{0.909}} & 23.88 & \textit{\textcolor{toptwo}{0.910}} & 23.63 \\
Dreambooth + AdaLoRA     & 0.898 & 23.74 & 0.911 & 21.82 & 0.917 & 25.39 & 0.907 & 23.93 & 0.905 & 23.67 \\
Dreambooth + SaRA        & 0.788 & \textbf{\textcolor{topone}{25.98}} & 0.884 & \textit{\textcolor{toptwo}{23.59}} & 0.887 & 25.35 & 0.882 & \textbf{\textcolor{topone}{25.51}} & 0.868 & \textit{\textcolor{toptwo}{24.99}} \\
 \textbf{Dreambooth + FeRA (Ours)} 
& \textbf{\textcolor{topone}{0.902}} & \textit{\textcolor{toptwo}{26.09}} 
& \textbf{\textcolor{topone}{0.923}} & \textbf{\textcolor{topone}{23.66}} 
& \textbf{\textcolor{topone}{0.925}} & \textbf{\textcolor{topone}{25.84}} 
& \textbf{\textcolor{topone}{0.909}} & \textit{\textcolor{toptwo}{24.50}} 
& \textbf{\textcolor{topone}{0.915}} & \textbf{\textcolor{topone}{25.36}} \\
\bottomrule
\end{tabular}%
}
\vspace{1mm}
\label{tab:clip_results_sd15}
\end{table*}

\vspace{-20mm}

\begin{table*}[b]
\centering
\caption{
Quantitative comparison between different PEFT methods on image customization (Stable Diffusion 3.0).
\textbf{\textcolor{topone}{Red bold}} = best, 
\textit{\textcolor{toptwo}{orange italic}} = second best.
}
\vspace{-2mm}
\small
\setlength{\tabcolsep}{4pt}
\renewcommand{\arraystretch}{1.12}
\resizebox{\linewidth}{!}{%
\begin{tabular}{l|cc|cc|cc|cc|cc}
\toprule
\multirow{2}{*}{Methods} &
\multicolumn{2}{c|}{Dog} &
\multicolumn{2}{c|}{Clock} &
\multicolumn{2}{c|}{Backpack} &
\multicolumn{2}{c|}{Toy Duck} &
\multicolumn{2}{c}{Teapot} \\
\cmidrule(lr){2-11}
 & CLIP-I~$\uparrow$ & CLIP-T~$\uparrow$
 & CLIP-I~$\uparrow$ & CLIP-T~$\uparrow$
 & CLIP-I~$\uparrow$ & CLIP-T~$\uparrow$
 & CLIP-I~$\uparrow$ & CLIP-T~$\uparrow$
 & CLIP-I~$\uparrow$ & CLIP-T~$\uparrow$ \\
\midrule
Dreambooth + Full-tuning & 0.798 & 24.55 & 0.799 & 23.55 & 0.664 & 24.49 & 0.800 & 24.45 & 0.760 & 24.52 \\
Dreambooth + LoRA       & 0.905 & 24.04 & 0.923 & 22.11 & \textit{\textcolor{toptwo}{0.927}} & 25.63 & 0.915 & 24.20 & 0.916 & 23.98 \\
Dreambooth + DoRA       & \textit{\textcolor{toptwo}{0.907}} & 24.11 & \textit{\textcolor{toptwo}{0.925}} & 22.18 & 0.924 & \textit{\textcolor{toptwo}{25.71}} & \textit{\textcolor{toptwo}{0.917}} & 24.28 & \textit{\textcolor{toptwo}{0.918}} & 24.05 \\
Dreambooth + AdaLoRA    & 0.906 & 24.09 & 0.924 & 22.16 & 0.927 & 25.69 & 0.916 & 24.25 & 0.917 & 24.03 \\
Dreambooth + SaRA       & 0.800 & \textbf{\textcolor{topone}{26.37}} & 0.897 & \textit{\textcolor{toptwo}{23.91}} & 0.896 & 25.67 & 0.895 & \textbf{\textcolor{topone}{25.90}} & 0.876 & \textit{\textcolor{toptwo}{25.52}} \\
 \textbf{Dreambooth + FeRA (Ours)} 
& \textbf{\textcolor{topone}{0.910}} & \textit{\textcolor{toptwo}{26.35}} 
& \textbf{\textcolor{topone}{0.930}} & \textbf{\textcolor{topone}{24.03}} 
& \textbf{\textcolor{topone}{0.935}} & \textbf{\textcolor{topone}{25.75}} 
& \textbf{\textcolor{topone}{0.920}} & \textit{\textcolor{toptwo}{25.00}} 
& \textbf{\textcolor{topone}{0.923}} & \textbf{\textcolor{topone}{25.72}} \\
\bottomrule
\end{tabular}%
}
\vspace{1mm}
\label{tab:clip_results_sd3}
\end{table*}

\newpage
\section{User Study (Human Evaluation)}
\label{user_study}
While automated metrics (e.g., FID, CLIP Score, MLLM Score) provide quantitative insights, they may not fully align with human perceptual preferences, especially for fine-grained style adaptation tasks. To rigorously assess the perceptual quality of FeRA, we conducted a User Study comparing our method against the strongest baseline, LoRA.

\textbf{Experimental Setup.} 
We randomly selected $M=30$ prompts from our test set, covering diverse styles. For each prompt, we generated image pairs using FeRA and LoRA with identical random seeds to ensure fairness. 
We invited $N=20$ evaluators to perform a blind Two-Alternative Forced Choice test. For each pair, evaluators were shown the reference style image and the two generated results (in randomized order) and asked to select the better one based on two criteria:
\begin{itemize}
    \item \textbf{Style Alignment:} Which image better captures the textures, brushstrokes, and color palette of the reference style?
    \item \textbf{Visual Fidelity:} Which image has better structural coherence and fewer artifacts?
\end{itemize}

\begin{figure}[htbp]
    \centering
    \includegraphics[width=0.75\linewidth]{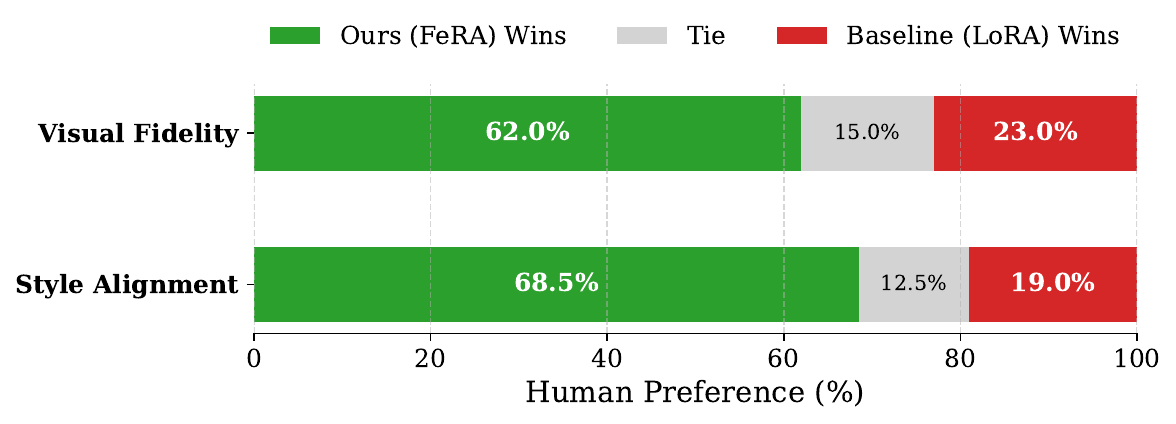}
    \caption{User study.}
    \label{fig:user_study}
\end{figure}

\textbf{Results.} 
Figure~\ref{fig:user_study} summarizes the voting results. FeRA significantly outperforms LoRA in human preference. Specifically, FeRA achieves a win rate of 68.5\% in Style Alignment and 62.0\% in Visual Fidelity. 

\end{document}